\documentclass[sigconf]{acmart}
\AtBeginDocument{%
  }

\copyrightyear{2025}
\acmYear{2025}
\setcopyright{cc}
\setcctype{by}
\acmConference[MM '25]{Proceedings of the 33rd ACM International Conference on Multimedia}{October 27--31, 2025}{Dublin, Ireland}
\acmBooktitle{Proceedings of the 33rd ACM International Conference on Multimedia (MM '25), October 27--31, 2025, Dublin, Ireland}\acmDOI{10.1145/3746027.3755755}
\acmISBN{979-8-4007-2035-2/2025/10}

\usepackage{listings}
\usepackage{verbatim}
\usepackage{indentfirst}
\usepackage[utf8]{inputenc}
\usepackage{framed}
\usepackage{rotating}
\def\mathbi#1{\textbf{\em #1}}

\usepackage{mathrsfs}
\usepackage{amsfonts}
\usepackage{mathtools}
\usepackage{algorithm}
\usepackage{algpseudocode}

\usepackage{pifont} 
\newcommand{\cmark}{\ding{51}} % Segno di spunta
\newcommand{\xmark}{\ding{55}} % Croce
\usepackage{adjustbox}
\usepackage{setspace}
\usepackage{booktabs}
\usepackage{longtable}
\usepackage{multirow}
\usepackage{pifont} 
\usepackage{underscore}
\usepackage[all]{hypcap}
\usepackage{makecell}
\usepackage{array}
\newcolumntype{H}{>{\setbox0=\hbox\bgroup}c<{\egroup}@{}}

\newcommand{\etal}{\textit{et al.~}}

\acmSubmissionID{5750}

\begin{document}

\title{Immunizing Images from Text to Image Editing via Adversarial Cross-Attention}

\author{Matteo Trippodo}
\email{matteo.trippodo@edu.unifi.it}
\affiliation{%
  \institution{University of Florence}
  \city{Florence}
  \country{Italy}
}
\author{Federico Becattini}
\email{federico.becattini@unisi.it}
\affiliation{%
  \institution{University of Siena}
  \city{Siena}
  \country{Italy}
}

\author{Lorenzo Seidenari}
\email{lorenzo.seidenari@unifi.it}
\affiliation{%
  \institution{University of Florence}
  \city{Florence}
  \country{Italy}
}

\renewcommand{\shortauthors}{Matteo Trippodo, Federico Becattini, and Lorenzo Seidenari}

\newcommand{\todo}[1]{{\color{red}\textbf{TODO}:  #1}}

\begin{abstract}
Recent advances in text-based image editing have enabled fine-grained manipulation of visual content guided by natural language. However, such methods are susceptible to adversarial attacks. In this work, we propose a novel attack that targets the visual component of editing methods.
We introduce Attention Attack, which disrupts the cross-attention between a textual prompt and the visual representation of the image by using an automatically generated caption of the source image as a proxy for the edit prompt. This breaks the alignment between the contents of the image and their textual description, without requiring knowledge of the editing method or the editing prompt.
Reflecting on the reliability of existing metrics for immunization success, we propose two novel evaluation strategies: Caption Similarity, which quantifies semantic consistency between original and adversarial edits, and semantic Intersection over Union (IoU), which measures spatial layout disruption via segmentation masks. Experiments conducted on the TEDBench++ benchmark demonstrate that our attack significantly degrades editing performance while remaining imperceptible.
\end{abstract}




%\centering
%  \includegraphics[width=0.75\textwidth]{images/AdversarialEditing_Eye.pdf}
 % seats. Ichiro Suzuki preparing to bat.}


\maketitle

\section{Introduction}

In the last few years, we observed a dramatic improvement in the quality of prompt-based image generative models, with Latent Diffusion~\cite{rombach2022high} being the most common and effective approach. The high availability of such models enabled new media production pipelines, empowering creators and streamlining several processes. Having generated images of such high quality and in certain cases hard to distinguish from natural ones, many abuse concerns have also been raised. A lot of effort is spent to align model output to ethical standards, removing offensive or in general undesired concepts from the models~\cite{DBLP:journals/corr/abs-2405-03486,qu2023unsafe,quaye2024adversarial}.
An even more concerning issue regards the capability of such models to edit existing images. Using easily available text-to-image editing pipelines, malicious users could download publicly available images and perform harmful edits to the image just by specifying concepts to add or remove via natural language.

\begin{figure}[!t]
    \centering
    \includegraphics[width=0.95\linewidth]{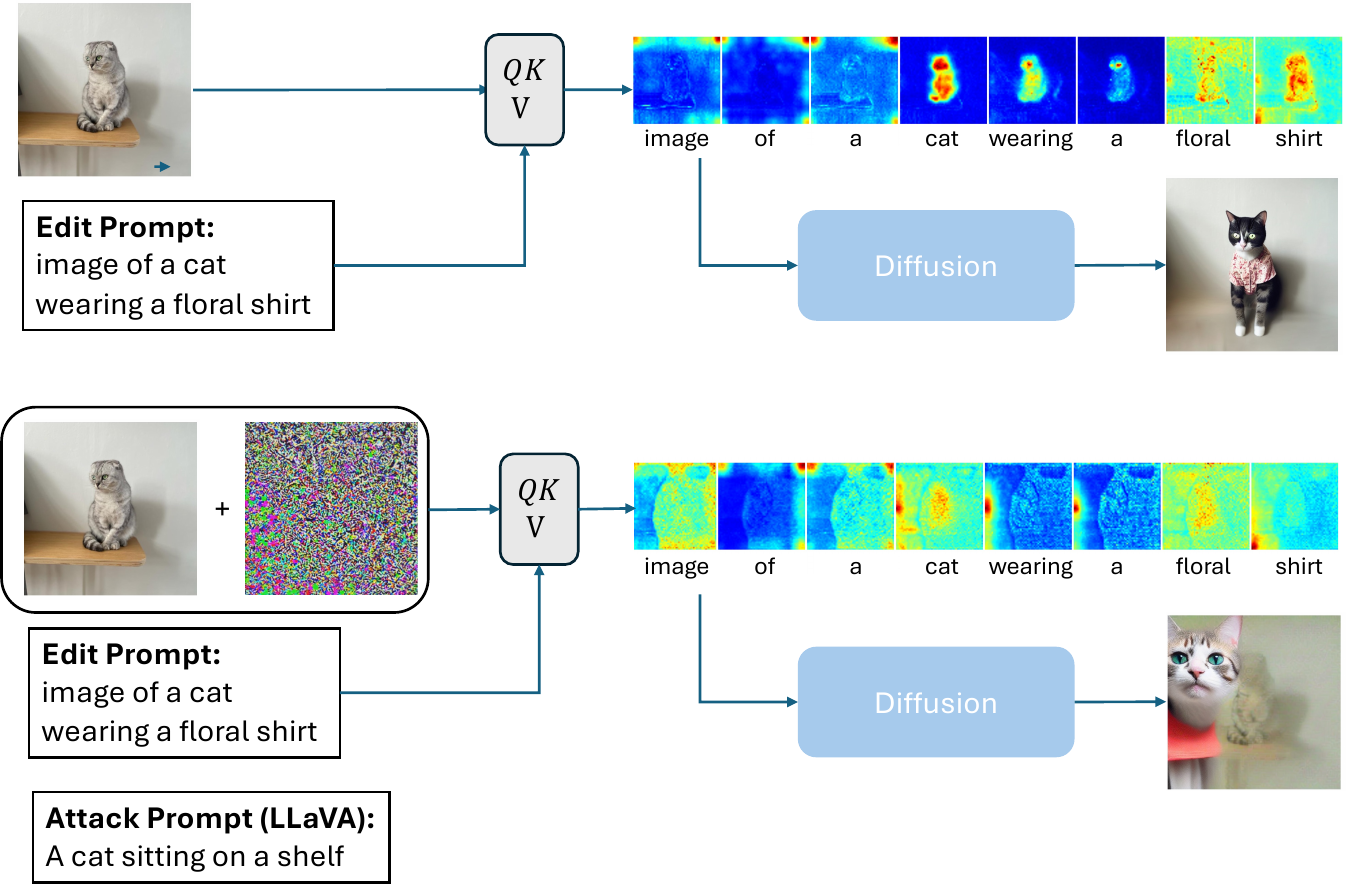}
    \caption{Adversarial noise is crafted to corrupt cross-attention. We craft an adversarial pattern to maximize the distance of attention from a surrogate prompt obtained by captioning the original image with LLaVa. Disruption of attention distribution of relevant tokens causes a failed edit.}
    \label{fig:eyecatcher}
\end{figure}

In this paper, we address this concern and propose a novel solution to defend against unwanted image edits. To immunize an image from an unwanted edit, we resort to adversarial machine learning. Adversarial machine learning is a fairly large sub-field of machine learning dealing with model vulnerabilities. Adversarial samples can be crafted to poison a training dataset preventing fine-tuning~\cite{fowl2021adversarial} or to produce incorrect model outputs~\cite{dong2020benchmarking}. 

In our case, we craft an adversarial noise to damage the image editing process. However, while for classification or regression tasks it is fairly easy to define failure cases, and hence devise an attack pipeline, for image editing measuring the success of an edit is harder and may in some cases be subjective.
Rather than reversing decisions based on the sample's category or regression target, prompt-based image editing is made possible by the computation of a cross-attention between the edit prompt and the input image feature. We argue that this is the main mechanism that must be disrupted to cause the largest damage. Attention serves two purposes: on the one hand, it represents the main communication path between the concept to add in the image and the low-level visual latent representation; on the other hand, it enables complex compositionality ensuring a coherent layout, allowing the model to add and remove concepts to an image without significantly altering the existing content.
One major impediment in attacking attention is represented by the lack of knowledge of the edit prompt. When protecting an image we can not know in advance what prompts will be used to edit it. We show that this issue can be overcome by using a surrogate edit prompt represented by an image caption. This will extract all the relevant tokens and damaging the cross-attention pattern of such tokens is enough to make the edit fail.
As a consequence, differently from other attacks, which mostly produce noisy or blurry results, our attention-based attack makes final images spatially and semantically inconsistent. Fig.~\ref{fig:eyecatcher} shows an example of our proposed approach.

To evaluate the success and unsuccess of image editing methods, we argue that existing metrics are insufficient. In fact, they are either focused on low-level image properties, such as SSIM, PSNR and LPIPS~\cite{zhang2018unreasonable}, or are too high-level like CLIP score~\cite{hessel2021clipscore}. To overcome these limitations, we propose two novel ways of evaluating edit quality, one relying on semantic segmentation to measure changes in the spatial layout of objects and one based on image captioning, measuring semantic inconsistencies.
In summary, the main contributions of our paper are the following.
\begin{itemize}
    \item We propose a novel attention-based immunization attack against image editing methods. By disrupting correspondences in the latent space between visual features and textual tokens, we effectively damage the editing pipeline.
    \item Our attack alters the spatial layout of the edited image, resulting in edits that deviate from the input image and cause unintended changes. The proposed attack also generates less visible adversarial artifacts compared to the competitors.
    \item We propose two novel metrics for image editing, Caption Similarity and Semantic Intersection over Union. Results on the TedBench++ dataset using three different editing models demonstrate both the effectiveness of our attack as well as the benefits of the proposed metrics.
\end{itemize}

\section{Related Works}
%Here we briefly cover Latent Diffusion Models and existing methods to defend against image editing.

%\subsection{Adversarial Machine Learning}

Adversarial machine learning comprises a set of techniques that aim at damaging the behavior of a targeted model. These adversarial examples can be imperceptibly different from legitimate inputs but can cause models to produce drastically different outputs \cite{PGDpaper,FGSMpaper,szegedy2014}. Generally speaking, an additive threat model is considered:  given a source image $x$, an adversarial samples $I_{adv} = I + \delta$, s.t. $||\delta||_{\infty}<\varepsilon$ is crafted, where the adversarial noise $\delta$ is obtained by ascending the model loss with a single step \cite{FGSMpaper}  or iteratively, as in Projected Gradient Descent (PGD)~\cite{PGDpaper}. In order to make the attack undetectable for the human eye, the norm of the noise is bounded by a perturbation budget ${\varepsilon}$.

%\subsection{Image Immunization}
Recently, with the exponential growth and availability of diffusion based models~\cite{croitoru2023diffusion,LDMpaper,sohl2015deep,dhariwal2021diffusion}, attacks to text-to-image editing methods are gaining popularity. These approaches aim at protecting a single image from potential text-guided edits performed by a diffusion model~\cite{salman2023raising,shan2023glaze,liang2023mist,basu2023editval,lin2024text}.
Immunization techniques are not always directed to diffusion models. Earlier works in fact were also devised to protect contents from manipulations with GANS~\cite{aneja2022tafim, yeh2021attack}.
As diffusion models are the de facto state of the art approach for image editing, we limit our analysis to this class of models. Diffusion model architectures can be roughly partitioned in three blocks: the encoder, the decoder and the noise prediction network, often implemented as a U-Net~\cite{ronneberger2015u}. Existing attacks target either the encoder $E(I)$ or the U-Net $\epsilon(E(I), t)$, with the latter attack being more expensive and requiring to iterate over multiple diffusion steps.

Salman \etal \cite{salman2023raising} present two attacks, one on the encoder and one on the diffusion process to push the final image towards a target gray image.
A similar strategy is proposed by Liang \etal \cite{liang2023mist} but use a watermark rather than a gray image to increase the disturbance in the final result. These works, although effective, require a predefined target to perform the attack. Its effectiveness thus may depend on such choice.
Other works instead find adversarial perturbations maximizing the $L_2$ norm of the initial predicted noise \cite{choi2024diffusionguard} or attack the VAE by inducing a posterior collapse using the KL Divergence as a loss \cite{guo2024grey}. Ozden \etal \cite{ozden2024optimization} instead protect images from specific edits altering the background of an image in an end-to-end fashion.

Other types of attacks for immunization have been presented in the literature, for instance by injecting triggers that can be recognized by the encoder~\cite{zeng2025guarddoor}. This kind of approach though requires to fine-tune the encoder of the diffusion model, limiting its applicability. Choi \etal~\cite{choi2024diffusionguard} instead use masks to protect parts of an image, making the editing process focus on sensitive regions, e.g. a face, impeding an effective alteration of the surrounding context. An immunization method targeted against DreamBooth models~\cite{ruiz2023dreambooth} was also proposed \cite{van2023anti} to defend against personalized text-to-image editings, i.e., generation processes targeted towards specific identities~\cite{gal2022image}.
Recently, Lo \etal \cite{Lo_2024_CVPR} proposed to manually pick a single concept to immunize in the image and generate a noise to suppress the cross-attention of a single token guided by a binary mask.
In a similar fashion, we propose an attack that targets cross-attention but we do not require any prior knowledge on the content of the image. In fact, we leverage a vision-language model to obtain a caption of the image that we can use as proxy for the textual edit prompt. In this way, we also avoid the dependence from a predefined image pattern as done in prior works \cite{salman2023raising,liang2023mist}.
To further motivate the idea of targeting cross-attention, we point out that its importance for effective prompt-based image editing has been highlighted in recent literature~\cite{yang2023dynamic}.

\begin{figure*}
    \centering
    \includegraphics[width=0.85\linewidth]{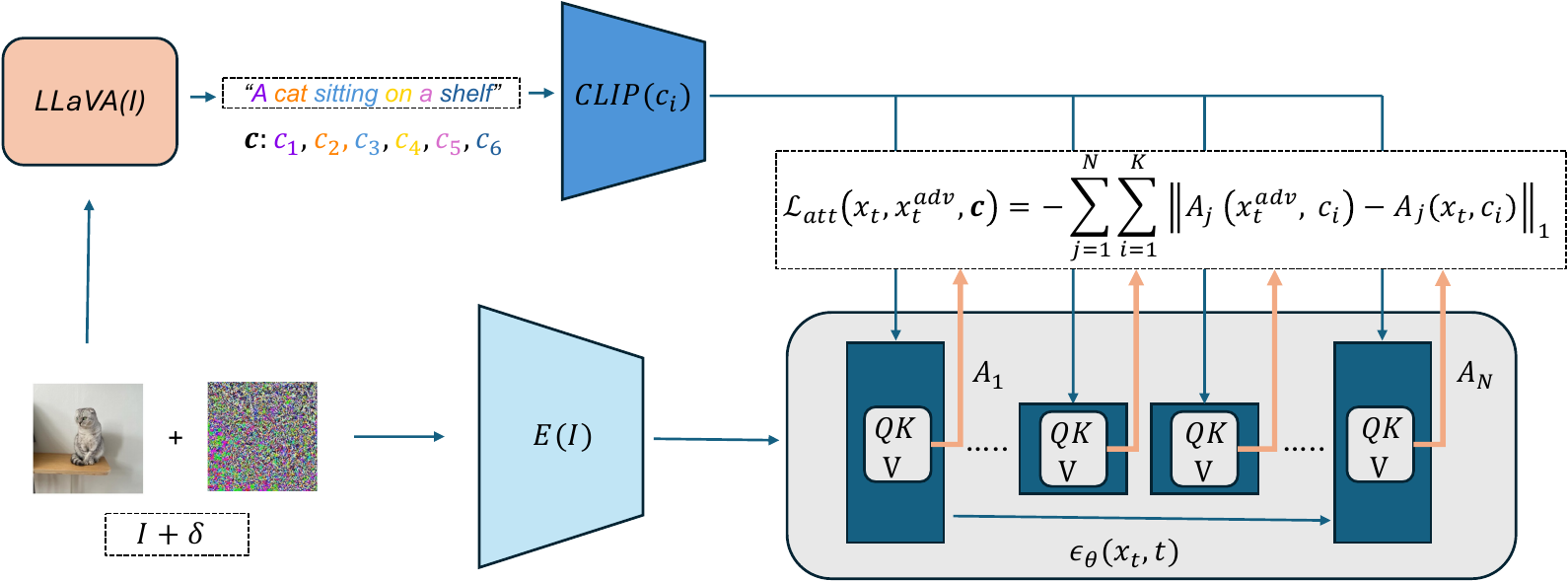}
    \caption{Overview of our framework. We craft $\delta$ to maximize the L1 distance between the cross-attention computed between latent embedding and a caption of the original image $c$. \label{fig:architecture}}
\end{figure*}
\section{Methodology}
Here we present our immunization strategy based on targeting cross-attention between text and the visual tokens. Since we focus on latent diffusion models, we first present a background on such architectures in Sec.~\ref{sec:ldm} to provide sufficient context to the reader and we then formally define the problem of image immunization against prompt-based editing methods in Sec.~\ref{sec:problem_def}. We present the details of our method in Sec.~\ref{sec:method}.

\subsection{Background on Latent Diffusion Models}
\label{sec:ldm}
Latent Diffusion Models (LDMs) are a class of generative models that improve the efficiency and scalability of diffusion-based image synthesis by operating in a compressed latent space rather than directly in the pixel space \cite{rombach2022high}. Instead of modeling the diffusion process in the high-dimensional image domain, LDMs first use a variational autoencoder (VAE) to encode images $I$ into a lower-dimensional latent representation $x = E(I)$. The diffusion process is then applied to this latent space, where it reverses noise after learning a noise prediction network $\epsilon(x_t, t)$ for each step $t \in [0, T]$~\cite{ho2020denoising}. 
After denoising, the latent representation is decoded back into the image space using the VAE decoder, $I_0=D(x_0)$. %This approach significantly reduces computational cost while maintaining high-quality generation, enabling applications such as text-to-image synthesis and image editing.

A cross-attention mechanism is leveraged to guide image generation. Specifically, during the denoising steps of the diffusion process, cross-attention layers integrate text embeddings (typically obtained from a pretrained language model such as CLIP~\cite{radford2021learning} or a Transformer encoder) with the latent visual features \cite{rombach2022high, ramesh2021zero}.

Formally, given a set of visual query features $Q \in \mathbb{R}^{n_q \times d}$ and textual key-value pairs $K, V \in \mathbb{R}^{n_k \times d}$, cross-attention computes
%\begin{equation}
    $\text{Attention}(Q, K, V) = \text{softmax}\left(\frac{QK^\top}{\sqrt{d}}\right)V.$
%\end{equation}
This allows the model to selectively focus on relevant parts of the text while generating or refining different regions of the image. Such conditioning ensures semantic coherence between the prompt and the generated image, enabling fine-grained control over attributes described in the text.

\subsection{Problem Definition}
\label{sec:problem_def}
We address image immunization against prompt-based editing with diffusion models. Prompt-based image editing can be formalized as follows. Given an original image $I_o$ and an edit prompt $\tau$ we assume to know an image editing model $\mathcal{M}(I_o, \tau)$ able to edit $I_o$ according to prompt $\tau$, obtaining an edited image $I_e$. 

The goal of immunization is to make $I_o$ \textit{difficult to edit}, resulting in a corrupted output of $\mathcal{M}(I_o, \tau)$. Image editing can fail in various ways: $I_e$ may be semantically distant from the intended prompt, may contain artifacts that make the quality of $I_e$ too low for its use, or the edit may simply not alter $I_o$. There is no clear way to define a successful or unsuccessful edit apart from using human evaluation. In \cite{brack2024ledits++}, metrics such as CLIP score and LPIPS are employed, alongside an evaluation where users provide feedback on the quality of the edits.
We aim at creating an adversarial noise $\delta, \text{s.t.}~  ||\delta||_{\infty}<\varepsilon$, with $\varepsilon$ being the attack budget that bounds the amount of alteration that can be done on the image. The generated noise, when added to the source image, will immunize it against diffusion-based editing, yielding incorrect or low-quality results.
For a comprehensive discussion on how to evaluate such edits, please refer to Sec.~\ref{sec:metrics}, where we also introduce new metrics for the task.

\subsection{Attention Based Immunization}
\label{sec:method}
Text-to-image diffusion models leverage a cross-attention mechanism to condition the diffusion process. Latent diffusion prompt-based methods are no different and require a cross-attention computation to correctly alter the image minimizing artifacts and preserving existing elements of the picture that should not undergo editing.

For this reason, we propose to inhibit cross-attention. Since we can not assume the edit prompt is known, we first extract a textual description of the image using the visual-language model LLaVa 1.5~\cite{liu2023visual}, using the prompt: \textit{"Give me a short caption to describe the main subject of this image. Use just few simple words"}.
Since edit prompts will likely address the reference in the image, we leverage the obtained caption as a proxy text that we can use as target for the attack.

Our attention-based attack is shown in Fig.~\ref{fig:architecture}. It consists in moving away the attention maps of each textual token with the adversarial image from the attention with the original image.
To this end, we optimize the loss $\mathcal{L}_{att}({x}_t, {x}^{adv}_t, \mathbi{c})$, between the latent representation of the image at step $t$ of the diffusion process, ${x}_t$, and its perturbed version ${x}^{adv}_t$ in function of the caption $\mathbi{c}$:

%\begin{equation}
%    \mathcal{L}_{att}(\mathbi{x}_t, \mathbi{x}_t + \delta, \mathbi{c}) =  - \sum_{c_i \in \mathbi{c}} \bigr\|A(\mathbi{x}_t + \delta, c_i) - A(\mathbi{x}_t,c_i) \bigr\|_1
%\end{equation}

\begin{equation}
    \mathcal{L}_{att}({x}_t, {x}^{adv}_t, \mathbi{c}) =  - \sum_{j=1}^N \sum_{i=1}^K \bigr\|A_j( {x}^{adv}_t), c_i) - A_j({x}_t,c_i) \bigr\|_1
\end{equation}

where, $A_j(\cdot, \cdot)$ is the cross-attention function a the j-th layer of $\epsilon$.
The loss, when minimized, maximizes the norm of the difference between the two attention maps, breaking the correlation between visual patterns and their corresponding textual tokens.

To perform our attack, we target the cross-attention in $T=10$ equally spaced timesteps $S_{T}=\{t_1, t,_2, ..., t_T\}$ in the diffusion process of Stable Diffusion. Therefore, we obtain the adversarial perturbation $\delta_{att}$ by optimizing the following with Projected Gradient Descent (PGD)~\cite{PGDpaper}:

\begin{equation}
    \delta_{att} \  = \text{arg} \min_{||\delta||_{\infty} \leq \varepsilon}  
     \mathcal{L}_{att}({x}_t, {x}^{adv}_t, \mathbi{c}).
\end{equation}

Algorithm~\ref{alg:attention_attack} illustrates the overall optimization process. In all our experiments, we use a budget $\varepsilon=16/255$ and number of iterations equal to $200$, with a step size of $2/255$.

    \begin{algorithm}[t]
\caption{Attention attack}\label{alg:attention_attack}
\begin{algorithmic}[1]
    \Require image $I$, caption $\mathbf{c} = \{c_1,c_2,...,c_K\}$, budget $\varepsilon$, step size $\alpha$, attack iterations $N$, diffusion timestep set $S_T = \{t_1, t_2, ..., t_T\}$
    \State $I^{adv} \gets I, \ \delta \gets 0$
    \State Encode the image: $x_0 \gets E(I)$
    \For {$t$ in $S_T$}
        %\State Add diffusion noise to the original image: $x^t \gets x$ 
        \State Diffuse at  timestep t: $x_t$
        %\State $x^t \gets$ Add diffusion noise to the original image $x$
        \State Store attentions between latent $x_t$ and $c$: $\{A_j(x_t,c_i)\}$
    \EndFor
    \For{$n = 1$ to $N$}
        %\State $t_i \sim \mathbf{U}(\{t_n\}_{n=1,...,T})$
        \State Sample a diffusion timestep: $t \sim \mathbf{U}(S_{T})$
        %\State Add diffusion noise to the immunized image: $x^t_{adv} \gets x_{adv}$
        \State Update adversarial image: $I^{adv} \gets I^{adv} + \delta$
        \State Encode the image: $x^{adv} \gets E(I^{adv})$
        \State Diffuse at timestep t: $x^{adv}_t$

        \State Compute $\mathcal{L}_{att} = 
        - \sum_{j=1}^N\sum_{i=1}^K ||A_j(x^{adv}_t, c_i) - A_j(x_t,c_i) ||_1$

        \State Update the adversarial perturbation:
        \Statex \hspace{\algorithmicindent} $\delta = \delta + \alpha \cdot \text{sign}(\nabla_{x_{adv}} \mathcal{L}_{att})$
        \Statex \hspace{\algorithmicindent} $\delta \gets \text{clip}(\delta, -\varepsilon, +\varepsilon)$
    \EndFor
    \Ensure Adversarial image $I^{adv}$
     \end{algorithmic}
    
    \end{algorithm}

\begin{table*}[t]
\centering
\small
\begin{adjustbox}{width=.90\textwidth,center}
\begin{tabular}{cccccccccccc}
\toprule
\textbf{8} & \textbf{16} & \textbf{32} & \textbf{64} & 
\textbf{LPIPS}$\uparrow$ & \textbf{PSNR}$\downarrow$ &  \textbf{SSIM} $\downarrow$ &  \textbf{Caption Sim.}$\downarrow$ & \textbf{oIoU} $\downarrow$ & \textbf{pIoU} $\downarrow$ \\
\midrule
% Righe pure
\cmark & \xmark & \xmark & \xmark & 0.487 ± 0.126 & 15.855 ± 2.837 & 0.609 ± 0.104 & 0.750 ± 0.159 & 0.534 ± 0.208 & 0.336 ± 0.248 \\ % attention8
\xmark & \cmark & \xmark & \xmark & 0.529 ± 0.158 & 15.795 ± 3.113 & 0.584 ± 0.107 & 0.745 ± 0.146 & 0.463 ± 0.211 & 0.274 ± 0.213 \\ % attention16
\xmark & \xmark & \cmark & \xmark & \underline{0.545 ± 0.150} & 15.818 ± 3.222 & \underline{0.560 ± 0.100} & 0.737 ± 0.161 & \underline{0.440 ± 0.208} & \underline{0.274 ± 0.200} \\ % attention32
\xmark & \xmark & \xmark & \cmark & 0.413 ± 0.091 & 18.047 ± 3.211 & 0.649 ± 0.092 & 0.818 ± 0.149 & 0.647 ± 0.180 & 0.385 ± 0.196 \\ % attention64
\midrule
% Combinazioni a due elementi
\cmark & \cmark & \xmark & \xmark & 0.508 ± 0.115 & 15.672 ± 2.641 & 0.596 ± 0.101 & 0.708 ± 0.129 & 0.466 ± 0.197 & 0.288 ± 0.200 \\ % attention8+16
\cmark & \xmark & \cmark & \xmark & 0.511 ± 0.134 & 15.750 ± 2.803 & 0.595 ± 0.105 & 0.719 ± 0.149 & 0.496 ± 0.193 & 0.290 ± 0.212 \\ % attention8+32
\cmark & \xmark & \xmark & \cmark & 0.514 ± 0.114 & 15.644 ± 2.729 & 0.595 ± 0.105 & 0.718 ± 0.151 & 0.501 ± 0.197 & 0.297 ± 0.189 \\ % attention8+64
%\xmark & \cmark & \cmark & \xmark & \textbf{0.560 ± 0.135} & \textbf{15.201 ± 2.887} & \textbf{0.558 ± 0.105} & 0.716 ± 0.186 & \underline{0.444 ± 0.200} & \textbf{0.253 ± 0.216} \\ % attention16+32
\xmark & \cmark & \cmark & \xmark & \textbf{0.570 ± 0.139}  & \textbf{15.173 ± 2.825} & \textbf{0.549 ± 0.097} & 0.727 ± 0.142   & \textbf{0.411 ± 0.195} & \textbf{0.238 ± 0.172} \\
\xmark & \cmark & \xmark & \cmark & 0.516 ± 0.138 & 15.756 ± 3.056 & 0.589 ± 0.097 & 0.752 ± 0.148 & 0.486 ± 0.192 & 0.303 ± 0.216 \\ % attention16+64
\xmark & \xmark & \cmark & \cmark & 0.461 ± 0.122 & 16.838 ± 2.989 & 0.614 ± 0.085 & 0.758 ± 0.153 & 0.532 ± 0.179 & 0.344 ± 0.218 \\ % attention32+64
\midrule
% Combinazioni a tre elementi
\cmark & \cmark & \cmark & \xmark & 0.509 ± 0.126 & 15.425 ± 2.726 & 0.598 ± 0.102 & \textbf{0.706 ± 0.161} & 0.495 ± 0.180 & 0.293 ± 0.202 \\ % attention8+16+32
\cmark & \cmark & \xmark & \cmark & 0.515 ± 0.112 & 15.473 ± 2.686 & 0.594 ± 0.094 & \underline{0.707 ± 0.140} & 0.468 ± 0.184 & 0.289 ± 0.195 \\ % attention8+16+64
\cmark & \xmark & \cmark & \cmark & 0.513 ± 0.111 & 15.692 ± 2.736 & 0.596 ± 0.097 & 0.731 ± 0.137 & 0.533 ± 0.204 & 0.306 ± 0.194 \\ % attention8+32+64
\xmark & \cmark & \cmark & \cmark & 0.542 ± 0.143 & 15.497 ± 3.058 & 0.572 ± 0.092 & 0.707 ± 0.165 & 0.457 ± 0.200 & 0.283 ± 0.179 \\ % attention16+32+64
\midrule
% Combinazione a quattro elementi
\cmark & \cmark & \cmark & \cmark & 0.530 ± 0.117 & \underline{15.321 ± 2.651} & 0.588 ± 0.092 & 0.752 ± 0.129 & 0.464 ± 0.178 & 0.287 ± 0.191 \\ % attention8+16+32+64
\bottomrule
\end{tabular}
\end{adjustbox}
\caption{Ablation study on cross-attention layer. Attack success metrics computed between original edit and adversarial edit.}
\label{tab:ablation-attention-2}
\end{table*}

\section{Experimental Results}
In the following, we discuss our experimental setting. We first introduce the dataset and present well-established as well as novel metrics in Sec.~\ref{sec:metrics}, then, in Sec.~\ref{sec:edit_methods} we present the edit methods that we defend against and in Sec.~\ref{sec:baselines} we discuss other immunization strategies. We follow by carrying out ablation studies discussing how to better implement our attack in Sec.~\ref{sec:ablation} and, finally, we report our results in Sec.~\ref{sec:results} and Sec.~\ref{sec:human_eval}, where we show comparisons with the state of the art and report a study with human evaluations.

\subsection{Dataset and Metrics}
\label{sec:metrics}

There is no universally accepted standard dataset for evaluating adversarial attacks on diffusion models, with each study using different image sets and prompts. For instance, Guo et al. \cite{guo2024grey} used 1,000 ImageNet images with five fixed prompts, ensuring variety in editing but limited semantic diversity;
Liang et al. \cite{AdvDM} used LSUN images but provided no details on the prompts, affecting reproducibility; Xue et al. \cite{diffprotect} used generic datasets like anime and WikiArt, without specifying the prompts; Lo et al. \cite{Lo_2024_CVPR} generated 150 artificial images but lacked detailed generation procedures.

In this paper, we use the TedBench++~\cite{brack2024ledits++} dataset for evaluating image editing techniques with diffusion models. The dataset includes 40 images and 127 editing prompts, covering a wide range of transformations such as style transfer, object removal, multiple conditioning, and complex substitutions.
We manually filtered the dataset to ensure reliable results with Stable Diffusion 1.5 \cite{LDMpaper}, selecting images where edits are successful using different editing methods, including SDEdit~\cite{SDEdit}, LEdits++~\cite{brack2024ledits++} and DDPM Inversion~\cite{editfriendlyddpmnoise}.

To assess the impact of adversarial perturbations, we used the commonly used PSNR, SSIM, LPIPS \cite{zhang2018unreasonable} and CLIP score \cite{hessel2021clipscore}. PSNR measures pixel-level differences but does not align with human perception, SSIM evaluates structural similarity and is more aligned with human visual quality than PSNR, while LPIPS estimates perceptual similarity using pre-trained neural networks and results to be more aligned with human perception.

These metrics, however, often fall short when assessing changes in the content or the spatial layout of objects. Indeed, LPIPS is not capable of detecting small semantic changes within the image.
%and, since it is based on a pre-trained model, the adversarial patterns in the images might deceive it.
On the other hand, SSIM by assessing structural similarity, often proves ineffective in capturing even substantial differences in terms of layout. 
CLIP score, instead, was introduced for image captioning and computes the similarity between a textual description and an image and is shown to align well with human judgment. However, if on the one hand it can provide a good way to align images and text, on the other hand, using CLIP score for different versions of the same image (e.g., the original and the edited one), will likely provide similar values as small details are not easily captured.

Therefore, to achieve a more comprehensive evaluation we propose two new metrics, \textit{Caption Similarity} and \textit{Semantic IoU}.
Caption Similarity evaluates the semantic similarity between images using captions generated by LLaVa 1.5\footnote{We use the prompt \textit{Give me a caption for this image}} and cosine similarity between textual embeddings. We extract textual embeddings using Sentence-BERT~\cite{reimers-2019-sentence-bert}\footnote{version all-MiniLM-L6-v2}, which provides 384-dimensional feature vectors.
More formally, the metric calculated for a pair of images $I_1$ and $I_2$ is defined as:
\begin{equation}
    \text{CaptionSimilarity} = \frac{\Phi(t_1)\Phi(t_2)}{\||\Phi(t_2)\||\||\Phi(t_2)\||}
\end{equation}

where $t_1$ and $t_2$ are the captions provided by LLaVa and $\Phi$ is the sentence encoder. This metric serves the purpose of assessing the semantic similarity of the contents of two images.

Semantic IoU instead measures changes in semantic layout. We first extract semantic segmentations from the two images using MaskFormer~\cite{cheng2021per} and we then define two variants of the metric, namely optimistic IoU (oIoU) and pessimistic IoU (pIoU).
In particular, referring to the predicted set of classes for the two images as $C_i$ e $C_j$, we have:

\textbf{Optimistic IoU} (\textit{oIoU}): considers only the intersection of the class labels predicted for the two images:
    \begin{equation}
        \text{oIoU}(I_1,I_2) = \frac{1}{|C_1 \cap C_2|}
        \sum_{c_i \in C_1 \cap C_2} \text{IoU}(M^{c_i}_1,M^{c_i}_2)
    \end{equation}
    where $M^{c_i}_1,M^{c_i}_2$ represent the binary masks $c_i$ relative to the images.

\textbf{Pessimistic IoU} (\textit{pIoU}): considers all the classes that appear in at last one of the two images:
    \begin{equation}
        \text{pIoU}(I_1,I_2) = \frac{1}{|C_1 \cup C_2|}
        \sum_{c_i \in C_1 \cup C_2} \text{IoU}(M^{c_i}_1,M^{c_i}_2).
    \end{equation}

\subsection{Editing Methods}
\label{sec:edit_methods}
We test our immunization on three text-to-image editing pipelines sharing the Stable Diffusion backbone~\cite{LDMpaper}.  
As a first edit method, we employed the image-to-image pipeline of Stable Diffusion, which is based on the SDEdit denoising process~\cite{SDEdit}. This is the simplest baseline, conditioning the image generation with an image and a prompt.\footnote{https://huggingface.co/docs/diffusers/en/api/pipelines/stable_diffusion/img2img}

As a more modern variation, we employed~\cite{editfriendlyddpmnoise}, which performs a DDPM inversion~\cite{ho2020denoising} designed to be edit-friendly, allowing to invert an image in the noise space in a way that allows its perfect reconstruction. With this method, after the inversion, the process can generate a new image which is conditioned by an edit prompt.

Finally, we use Ledits++~\cite{brack2024ledits++}, a state of the art editing method. Ledits++ combines a perfect inversion module based on the SDE solver from \cite{lu2022dpm} with a per-token guidance process inspired by classifier-free guidance and a mask computation, based on attention, to ground the editing. This method is the best performing on clean images and also the hardest to attack, as shown in Tab.\ref{tab:metriche-risultati-finali} and in Q1 in Tab.\ref{tab:human_tests}.

\subsection{Baselines}
\label{sec:baselines}
We compare our attack strategy with competitors from the state of the art, namely Posterior Collapse Attack (PCA)~\cite{guo2024grey} and PhotoGuard's encoder attack (PG)~\cite{salman2023raising}.

The Posterior Collapse Attack aims at inducing posterior collapse~\cite{razavi2019preventing}, a common phenomenon in variational autoencoders that occurs when latent variables are not able to effectively interpret the input data. To achieve this goal, the attack minimizes the KL divergence between the learned posterior distribution $q(z|x)$ and the target distribution $p^*(z)$. The adversarial noise is thus obtained by iteratively applying projected sign gradient ascent.
PhotoGuard's encoder attack instead targets the embedding of the image produced by the encoder of the latent diffusion model to make it close to the one of a pattern image. In the original formulation, the authors leverage a gray pattern, whereas here we use a text pattern as in MIST~\cite{liang2023mist}, to make the attack more disruptive. The adversarial noise is obtained iteratively through Projected Gradient Descent.

We also introduce an additional baseline that, similarly to PhotoGuard, disrupts the image embedding yielded by the encoder in the VAE. The purpose of this attack is to maximize the distance between the encoding of the original image and the one of the adversarial image. Differently from state of the art methods such as PhotoGuard and Mist \cite{liang2023mist}, which use a predefined pattern, this baseline does not require any image target. This makes the attack simpler, yet more versatile. We refer to this attack as \textit{Least Cosine Similarity Attack} (LCSA).

More formally, to craft the adversarial perturbation $\delta_{LCSA}$, the following loss function is optimized:

\begin{equation}
    \delta_{LCSA} = \text{arg} \min_{||\delta||_{\infty} \leq \varepsilon}  \ S_{cos}(E(\mathbi{I}), {E}(\mathbi{I}+\delta))
\end{equation}

where $S_{cos}$ is the cosine similarity and ${E}(\cdot)$ is the operator that extracts the latent representation of the input with the VAE encoder. In our experiments, given an input image $\mathbi{I}$, we have ${E}(I) = \mathbi{x} \in~\mathbb{R}^{4 \times 64 \times 64}$ as we use Stable Diffusion 1.5 \cite{LDMpaper}.

\begin{table}[!htb]
\centering
\small
\begin{adjustbox}{width=.9\columnwidth, center}
\begin{tabular}{ccccccc}
\toprule
\textbf{8} & \textbf{16} & \textbf{32} & \textbf{64} & \textbf{LPIPS} $\downarrow$ & \textbf{PSNR} $\uparrow$ & \textbf{SSIM} $\downarrow$ \\
\midrule
% Righe pure
\cmark & \xmark & \xmark & \xmark & 0.247 ± 0.089 & 33.387 ± 0.319 & 0.880 ± 0.039 \\ % attention8
\xmark & \cmark & \xmark & \xmark & 0.248 ± 0.083 & 33.250 ± 0.331 & \underline{0.878 ± 0.037} \\ % attention16
\xmark & \xmark & \cmark & \xmark & 0.253 ± 0.083 & 33.187 ± 0.404 & 0.879 ± 0.038 \\ % attention32
\xmark & \xmark & \xmark & \cmark & \textbf{0.241 ± 0.090} & \textbf{33.647 ± 0.300} & 0.884 ± 0.037 \\ % attention64
\midrule
% Combinazioni a due elementi
\cmark & \cmark & \xmark & \xmark & 0.248 ± 0.087 & 33.315 ± 0.347 & 0.881 ± 0.038 \\ % attention8+16
\cmark & \xmark & \cmark & \xmark & 0.246 ± 0.085 & 33.348 ± 0.343 & 0.881 ± 0.038 \\ % attention8+32
\cmark & \xmark & \xmark & \cmark & \underline{0.244 ± 0.086} & \underline{33.425 ± 0.308} & 0.882 ± 0.038 \\ % attention8+64
\xmark & \cmark & \cmark & \xmark & 0.247 ± 0.084 & 33.140 ± 0.333 & \textbf{0.877 ± 0.039} \\ % attention16+32
%\xmark & \cmark & \cmark & \xmark & 0.249 ± 0.080 & 32.994 ± 0.483 & 0.877 ± 0.037 \\
\xmark & \cmark & \xmark & \cmark & 0.249 ± 0.088 & 33.332 ± 0.339 & 0.881 ± 0.039 \\ % attention16+64
\xmark & \xmark & \cmark & \cmark & \underline{0.244 ± 0.087} & 33.424 ± 0.399 & 0.885 ± 0.037 \\ % attention32+64
\midrule
% Combinazioni a tre elementi
\cmark & \cmark & \cmark & \xmark & 0.246 ± 0.085 & 33.306 ± 0.351 & 0.881 ± 0.038 \\ % attention8+16+32
\cmark & \cmark & \xmark & \cmark & 0.247 ± 0.087 & 33.344 ± 0.359 & 0.882 ± 0.038 \\ % attention8+16+64
\cmark & \xmark & \cmark & \cmark & 0.245 ± 0.083 & 33.365 ± 0.365 & 0.881 ± 0.038 \\ % attention8+32+64
\xmark & \cmark & \cmark & \cmark & 0.246 ± 0.086 & 33.243 ± 0.338 & 0.880 ± 0.040 \\ % attention16+32+64
\midrule
% Combinazione a quattro elementi
\cmark & \cmark & \cmark & \cmark & 0.247 ± 0.085 & 33.312 ± 0.367 & 0.883 ± 0.036 \\ % attention8+16+32+64
\bottomrule
\end{tabular}
\end{adjustbox}
\caption{Ablation study on cross-attention layer. Imperceptibility metrics between original image and adversarial image.}\label{tab:ablation_perceptiveness}
%\vspace{-20pt}
\end{table}

\subsection{Ablation Study}
\label{sec:ablation}
We first carry out an ablation study to assess which layer to target with the attention attack. The Stable Diffusion 1.5 model employs cross-attention at different resolutions, yielding attention maps of spatial resolution $8 \times 8$, $16 \times 16$, $32 \times 32$ and $64 \times 64$.
To establish the impact of each layer on the effectiveness of the attack, we test the attack on each individual layer as well as on all possible combinations of layers.
In Tab.~\ref{tab:ablation-attention-2} and Tab.~\ref{tab:ablation_perceptiveness} we report our findings respectively on the effectiveness of the attack and its perceptibility, using the LEdits++~\cite{brack2024ledits++} editing method.

Tab.~\ref{tab:ablation_perceptiveness} reports LPIPS, PSNR and SSIM computed between the original and adversarial images. The attack involving only the $64 \times 64$ attention map provides the most imperceptible noise in terms of LPIPS and PSNR, whereas the best SSIM is obtained when attacking the combination of $16 \times 16$ and $32 \times 32$ maps. However, overall the metrics do not vary significantly for different configurations.

On the contrary, Tab.~\ref{tab:ablation-attention-2} shows that targeting different attention maps can indeed bring relevant improvements to the effectiveness of the attack.
The best overall configuration is the combination of $16 \times 16$ and $32 \times 32$ maps. An interpretation of this behavior is that attention maps at the lowest and highest resolutions capture details that are either too fine or too coarse. In fact, even when used alone, they do not provide satisfactory results, contrary to the intermediate-sized maps.
Interestingly, however, between the $8 \times 8$ and the $64 \times 64$ ones, attacking the coarser maps is better.
%Interestingly, targeting all the attention maps at once is detrimental to the attack.
The only metric that benefits from the combination of the other maps is Caption Similarity, for which the best results are obtained when attacking three different resolutions.

Based on these results, in the following, we will refer to Attention Attack as our proposed method targeting both the $16 \times 16$ and the $32 \times 32$ maps.

\begin{table}[t]
\centering
\small
\resizebox{0.9\columnwidth}{!}{
\begin{tabular}{lccc}
\toprule
\textbf{Attack type} & \textbf{LPIPS} ↓ & \textbf{PSNR} ↑ & \textbf{SSIM} $\uparrow$ \\
\midrule    
Posterior Collapse  & \underline{0.289 ± 0.083}  & \underline{31.833 ± 0.253}   & 0.847 ± 0.042  \\
PhotoGuard    & 0.290 ± 0.097  & 31.428 ± 0.176  & \underline{0.850 ± 0.043}  \\
%Attention   & \textbf{0.249 ± 0.080} & \textbf{32.994 ± 0.483} & \textbf{0.877 ± 0.037} \\
LCSA  & 0.358 ± 0.108 & 31.212 ± 0.189   & 0.808 ± 0.065  \\
Attention & \textbf{0.247 ± 0.084} & \textbf{33.140 ± 0.333} & \textbf{0.877 ± 0.039} \\
\bottomrule
\end{tabular}
}
\caption{Imperceptibility metrics between original image and adversarial image for different immunization methods.}
\label{tab:final-result-adv-quality}
\end{table}

\begin{table*}[ht]
\centering
\begin{adjustbox}{width=.9\textwidth, center}
\begin{tabular}{lccccccccc}
\toprule
\textbf{Attack Type} & \textbf{Editing Method} & \textbf{CLIP($I_o^e$,$\tau$)} $\uparrow$ & \textbf{CLIP($I_{adv}^e$,$\tau$)} $\downarrow$  & \textbf{LPIPS} $\uparrow$ & \textbf{PSNR} $\downarrow$ & \textbf{SSIM} $\downarrow$  & \textbf{Caption Sim.}$\downarrow$  & \textbf{oIoU} $\downarrow$  & \textbf{pIoU} $\downarrow$  \\
\midrule
% Blocco Img2Img
Posterior Collapse    & \multirow{4}{*}{SDEdit~\cite{SDEdit}} & \multirow{4}{*}{17.330 ± 2.305} & 17.375 ± 2.246 & 0.756 ± 0.082 & \textbf{9.383 ± 1.359}  & \underline{0.259 ± 0.133} & \textbf{0.697 ± 0.160}  & \underline{0.288 ± 0.172}  & \underline{0.129 ± 0.098}  \\
PhotoGuard     &    &    &   17.293 ± 2.281               &  0.704 ± 0.086 & \underline{9.625 ± 1.404} & 0.319 ± 0.131             & 0.710 ± 0.154            & 0.345 ± 0.194             & 0.156 ± 0.123             \\
LCSA      &   &   & \underline{17.276 ± 2.185}                 & \textbf{0.837 ± 0.118}  & 9.829 ± 1.378           & \textbf{0.164 ± 0.070}  & 0.721 ± 0.159           & 0.351 ± 0.187             & 0.158 ± 0.123             \\
Attention &      &     & \textbf{17.031 ± 2.295}            & \underline{0.793 ± 0.091}   & 9.686 ± 1.353           & \underline{0.259 ± 0.114}    & \underline{0.699 ± 0.173}    & \textbf{0.227 ± 0.125}      & \textbf{0.109 ± 0.086}      \\
\midrule
% Blocco LEdits++
Posterior Collapse    & \multirow{4}{*}{LEdits++~\cite{brack2024ledits++}} & \multirow{4}{*}{17.591 ± 1.911} & 17.597 ± 1.758  & \underline{0.599 ± 0.138} & \textbf{14.779 ± 2.097}  & 0.606 ± 0.118 & \underline{0.743 ± 0.149}  & 0.564 ± 0.169 & 0.313 ± 0.206 \\
PhotoGuard     &    &       & \underline{17.569 ± 1.835}                 & 0.443 ± 0.103  & 15.777 ± 2.636         & 0.641 ± 0.107 & 0.784 ± 0.162            & 0.594 ± 0.183 & 0.394 ± 0.224 \\
LCSA     &     &         & 17.584 ± 2.149              & \textbf{0.699 ± 0.145}  & 15.939 ± 2.385      & \textbf{0.362 ± 0.104}  & 0.801 ± 0.131           & \underline{0.551 ± 0.179}  & \underline{0.268 ± 0.171} \\
Attention &   &       & \textbf{17.560 ± 1.699}               & 0.570 ± 0.139  & \underline{15.173 ± 2.825} & \underline{0.549 ± 0.097} & \textbf{0.727 ± 0.142}   & \textbf{0.411 ± 0.195} & \textbf{0.238 ± 0.172} \\
\midrule
% Blocco ddpm
Posterior Collapse    & \multirow{4}{*}{\makecell{DDPM\\Inversion}~\cite{editfriendlyddpmnoise}} & \multirow{4}{*}{\makecell{17.717 ± 1.744}} & 17.697 ± 1.925 & \underline{0.619 ± 0.123} & \textbf{16.067 ± 1.552} & 0.560 ± 0.126  & \underline{0.738 ± 0.175} & 0.589 ± 0.183  & 0.296 ± 0.209  \\
PhotoGuard       &      &        & 17.623 ± 1.839             & 0.520 ± 0.112          & 16.278 ± 1.641         & 0.516 ± 0.111  & 0.750 ± 0.180          & \underline{0.545 ± 0.196} & 0.281 ± 0.213  \\
LCSA     &     &       & \underline{17.526 ± 1.753}               & \textbf{0.718 ± 0.142} & 16.169 ± 1.404         & \textbf{0.275 ± 0.085} & 0.739 ± 0.182          & 0.575 ± 0.205  & \underline{0.243 ± 0.143} \\
Attention&    &          & \textbf{17.508 ± 1.981}             & 0.614 ± 0.095          & \underline{16.133 ± 1.820}& \underline{0.466 ± 0.093} & \textbf{0.675 ± 0.174}  & \textbf{0.344 ± 0.212} & \textbf{0.165 ± 0.127} \\ %\midrule

\bottomrule
\end{tabular}
\end{adjustbox}
\caption{Immunization success rate in term of low-level and high-level metrics for three editing methods.}
\label{tab:metriche-risultati-finali}
\end{table*}

\begin{figure}[t]
    \centering
    \includegraphics[width=\columnwidth]{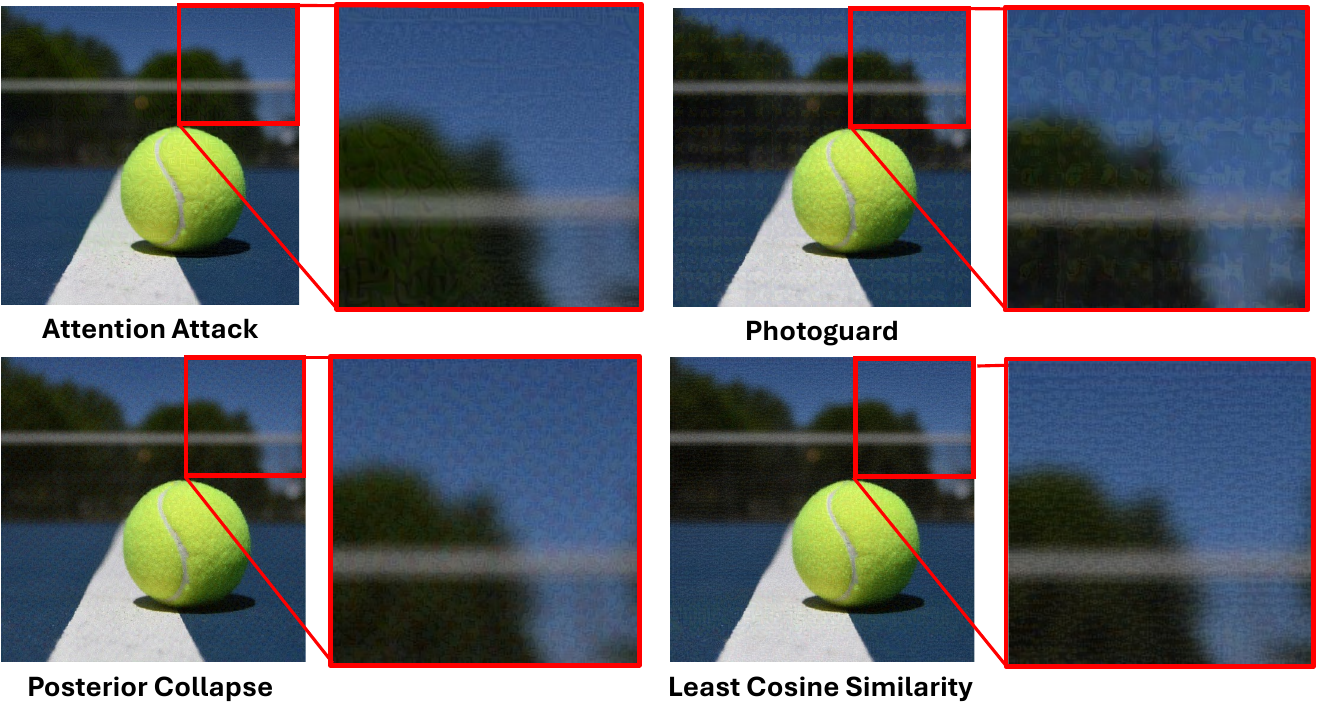}
    \caption{Noise patterns generated by different attacks. Attention Attack introduces the most imperceptible perturbations.}
    \label{fig:pattern_sample}
\end{figure}

\begin{table*}[t]
\centering
\resizebox{0.9\textwidth}{!}{
\begin{tabular}{lccc|ccc|ccc|ccc}
\toprule
& \multicolumn{3}{c}{\textbf{Q1: Edit Success $\downarrow	
$}} & \multicolumn{3}{c}{\textbf{Q2: Worst Edit $\uparrow	
$}} & \multicolumn{3}{c}{\textbf{Q3: Worst Quality $\uparrow	
$}} & \multicolumn{3}{c}{\textbf{Q4: Spatial Layout $\uparrow	
$}} \\ \hline
\textbf{Method} & \textbf{DDPMinv} & \textbf{LEdits} & \textbf{SDEdit} & \textbf{DDPMinv} & \textbf{LEdits} & \textbf{SDEdit} & \textbf{DDPMinv} & \textbf{LEdits} & \textbf{SDEdit} & \textbf{DDPMinv} & \textbf{LEdits} & \textbf{SDEdit} \\
\hline
Posterior Collapse & 0.379 & 0.404 & 0.464 & 0.119 & 0.149 & 0.189 & 0.080 & 0.185 & 0.196 & 0.296 & 0.243 & 0.557\\

PhotoGuard & 0.356 & 0.493 & 0.472 & 0.079 & 0.074 & 0.168 & 0.330 & 0.098 & 0.206 & 0.194 & 0.243 & 0.509\\

LCSA  & 0.477 & 0.456 & 0.472 & 0.089 & 0.106 & 0.200 & 0.080 & 0.109 & 0.168 & 0.122 & 0.122 & 0.330\\ 

Attention & \textbf{0.182} & \textbf{0.235}  & \textbf{0.248} & \textbf{0.525} & \textbf{0.436} & \textbf{0.242} & \textbf{0.500} & \textbf{0.554} & \textbf{0.280} & \textbf{0.765} & \textbf{0.635} & \textbf{0.755}\\ \hline

Original edit & 0.636 & 0.713 & 0.520 & 0.168 & 0.181 & 0.105 & 0.011 & 0.054 & 0.131 & 0.061 & 0.135 & 0.255\\
\bottomrule
\end{tabular}
}
\caption{Human Evaluation for four different questions. For all tests we report if higher ($\uparrow$) or lower ($\downarrow$) is better in terms of immunization. Our approach is consistently superior to other immunization approaches for all three editing methods.}
%\vspace{-10pt}
\label{tab:human_tests}
\end{table*}

\subsection{Comparison with the State of The Art}
\label{sec:results}
We now present a comparison against the state of the art, comparing our proposed approach against the baselines discussed in Sec.~\ref{sec:baselines}.
We first measure how imperceptible the attack is. In Tab.~\ref{tab:final-result-adv-quality} we compare LPIPS, PSNR and SSIM between the original images and their attacked version, for all immunization strategies.
Our attention attack yields the less visible attacks, which is also confirmed by Fig.~\ref{fig:pattern_sample}, in which we show the patterns obtained with different attacks. It must be noted that all the attacks are performed using the same budget $\varepsilon=16/255$.

Tab.~\ref{tab:metriche-risultati-finali} instead reports LPIPS, PSNR and SSIM between the edited original image and edited adversarial image. We compare our results against the state of the art for three editing methods, SDEdit~\cite{SDEdit}, LEdits++~\cite{brack2024ledits++} and DDPM Inversion~\cite{editfriendlyddpmnoise}. Looking at these metrics, it is hard to assess which immunization method is better, with the Least Cosine Similarity Attack obtaining good results for LPIPS and SSIM and Posterior Collapse for PSNR. Our attention attack instead, according to these metrics, performs slightly worse, often resulting the second-best. However, it can be seen that, despite there being variability across editing methods in terms of absolute values, the relative margin between the attacks is small and all the results exhibit a high variance. Furthermore, as previously discussed, we argue that these metrics, although commonly used, are not well-suited for assessing the quality of immunization methods against image editing. This is supported by our novel metrics discussed further and by human evaluation (see Sec~\ref{sec:human_eval}).

We also report the CLIP score between the edited adversarial image $I^e_{adv}$ and the editing prompt $\tau$. As a reference, for each editing method, we also report the CLIP score between the edited original image $I^e_o$ and the prompt. It can be seen that this metric is more stable across different methods compared to, LPIPS, PSNR and SSIM, yet it shows very small differences between the immunization strategies. Whereas the attention attack appears to consistently provide the lowest score, surprisingly in a few cases the CLIP score for the edited adversarial image results to be higher to the edited original one (e.g., Posterior Collapse with SDEdit and LEdits++). This points out the poor adequacy of the CLIP score metric for this task.

\begin{figure*}[t]
\centering
\resizebox{.9\textwidth}{!}{
\begin{tabular}{c|ccc|cccc}
\toprule
    \textbf{Editing Method} & \textbf{Original} & \textbf{Prompt} & \textbf{\makecell{Original\\Edit}} & \textbf{PhotoGuard} & \textbf{\makecell{Posterior\\Collapse}} & \textbf{LCSA} & \textbf{Attention} \\ \midrule
    \multirow[b]{3}{*}{\makecell{DDPM\\Inversion}~\cite{editfriendlyddpmnoise}} &
     \includegraphics[width=.09\textwidth]{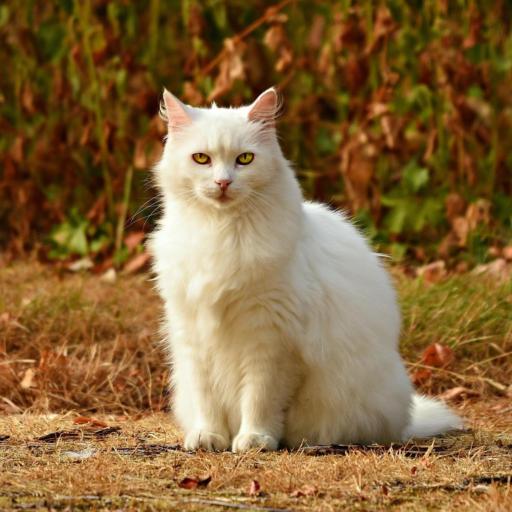} &
     \makecell[b]{A photo of a cat\\in a grass field.\\~} &
    \includegraphics[width=.09\textwidth]{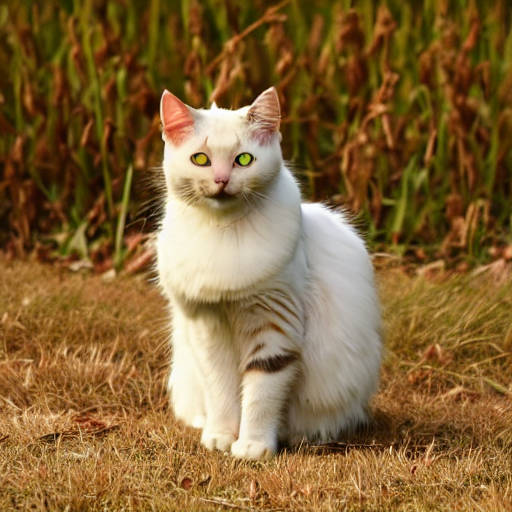} &
    \includegraphics[width=.09\textwidth]{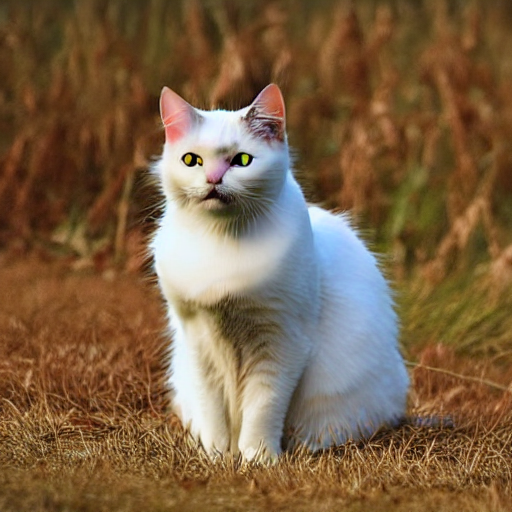} &
    \includegraphics[width=.09\textwidth]{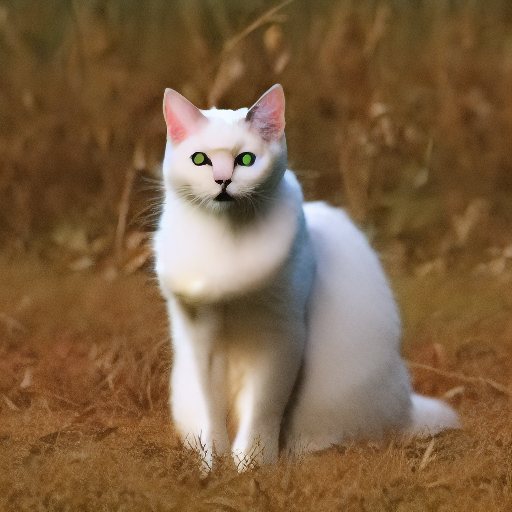} &
    \includegraphics[width=.09\textwidth]{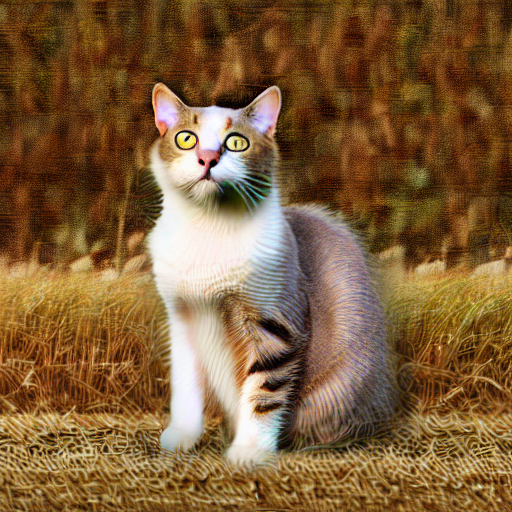} &
    \includegraphics[width=.09\textwidth]{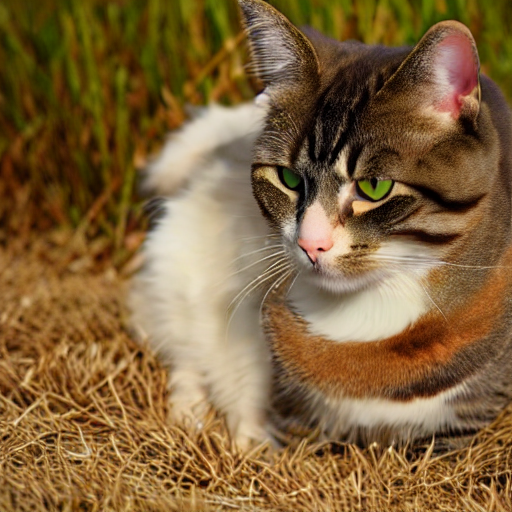} \\

     &
    \includegraphics[width=.09\textwidth]{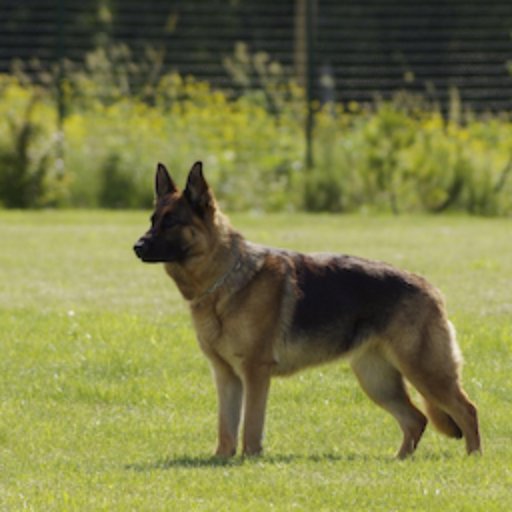} &
    \makecell[b]{A photo of\\a jumping dog.\\~} &
    \includegraphics[width=.09\textwidth]{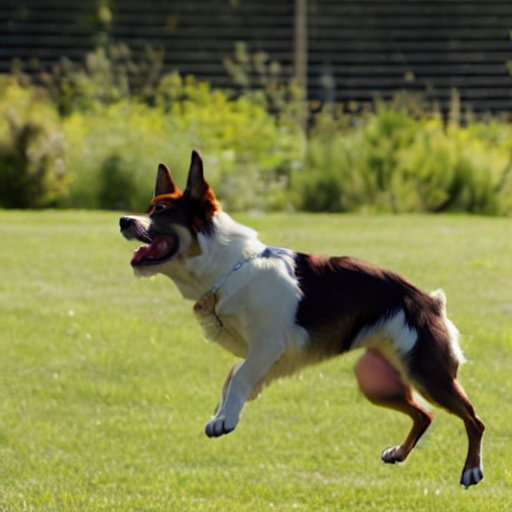} &
    \includegraphics[width=.09\textwidth]{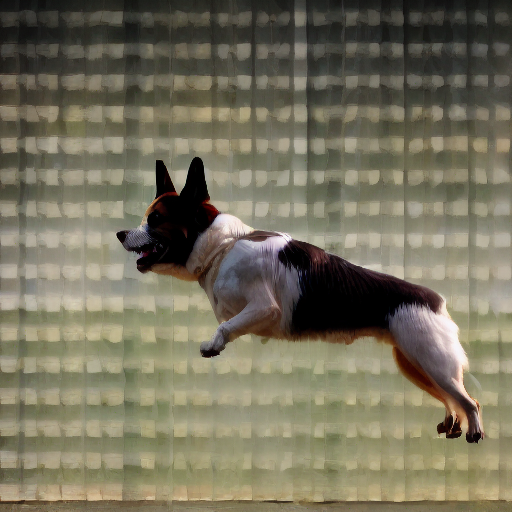} &
    \includegraphics[width=.09\textwidth]{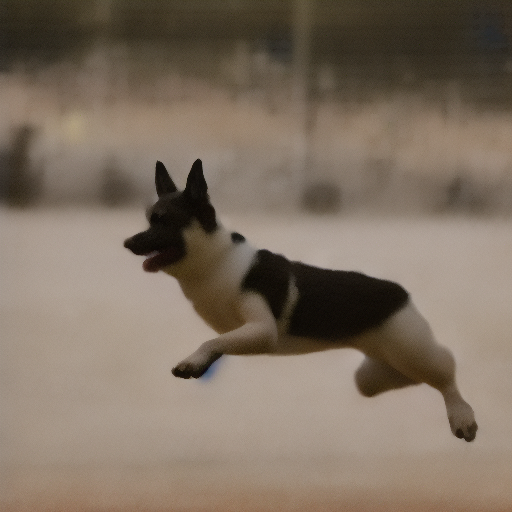} &
    \includegraphics[width=.09\textwidth]{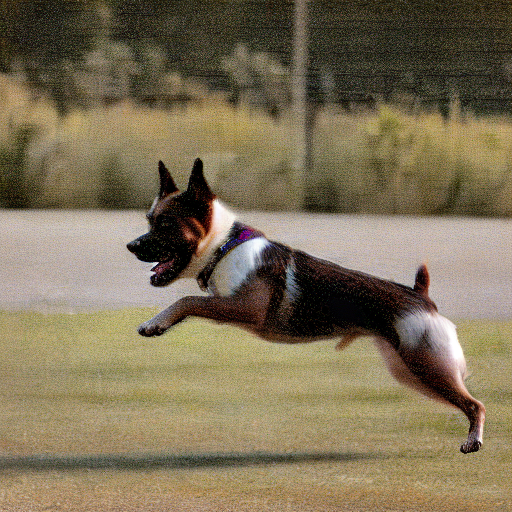} &
    \includegraphics[width=.09\textwidth]{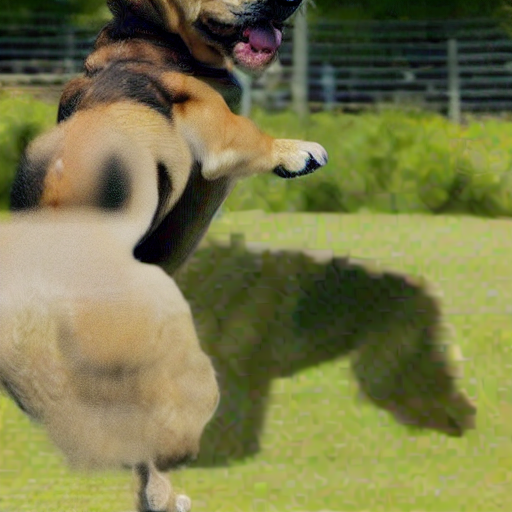} \\

     &
    \includegraphics[width=.09\textwidth]{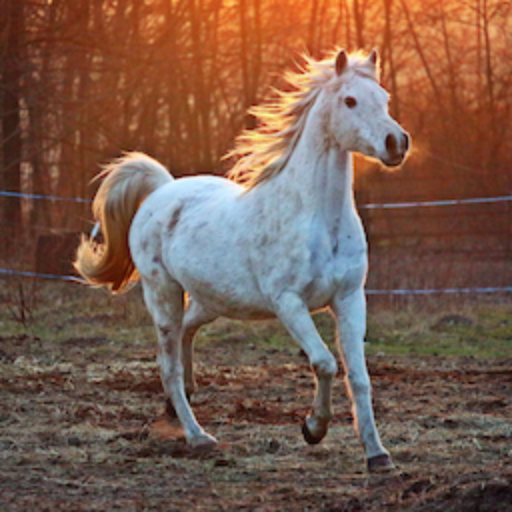} &
    \makecell[b]{A circus horse.\\~\\~} &
    \includegraphics[width=.09\textwidth]{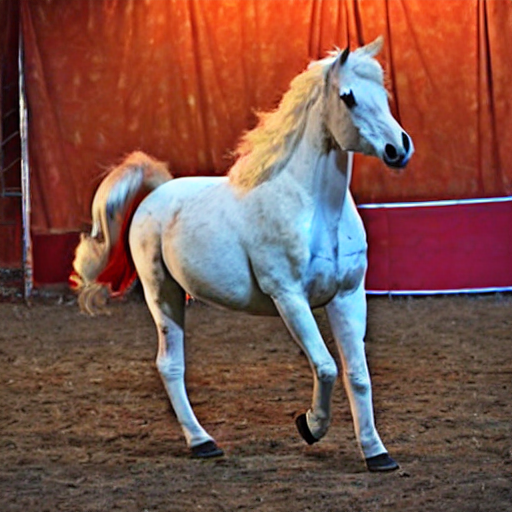} &
    \includegraphics[width=.09\textwidth]{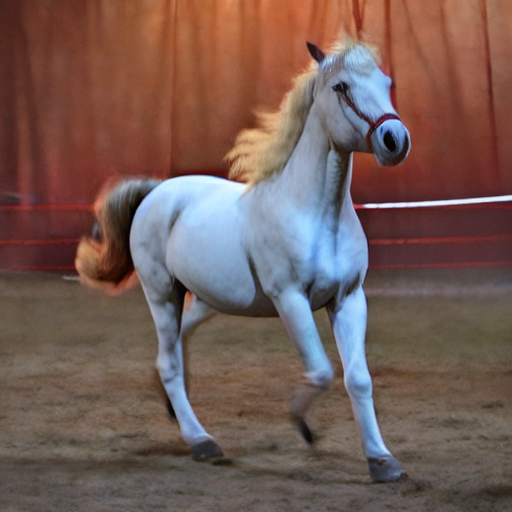} &
    \includegraphics[width=.09\textwidth]{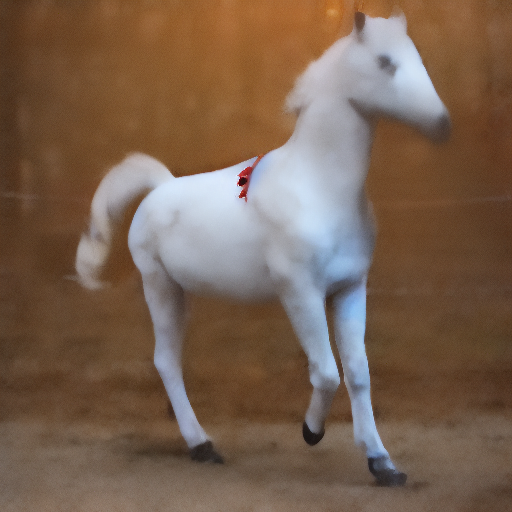} &
    \includegraphics[width=.09\textwidth]{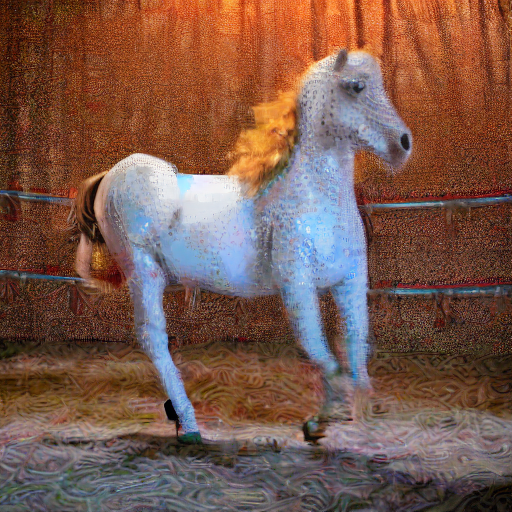} &
    \includegraphics[width=.09\textwidth]{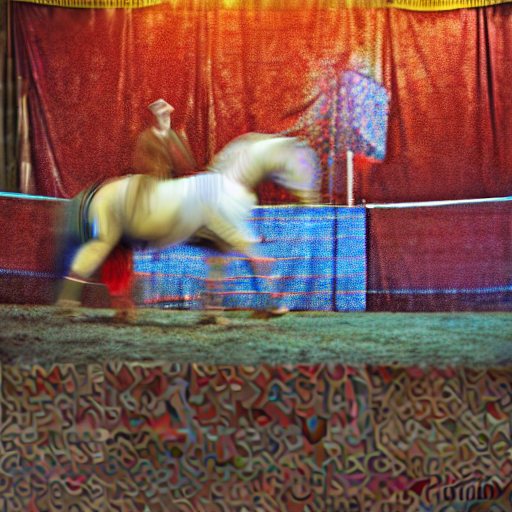} \\ \midrule

    \multirow[b]{3}{*}{LEdits++~\cite{brack2024ledits++}} &
    \includegraphics[width=.09\textwidth]{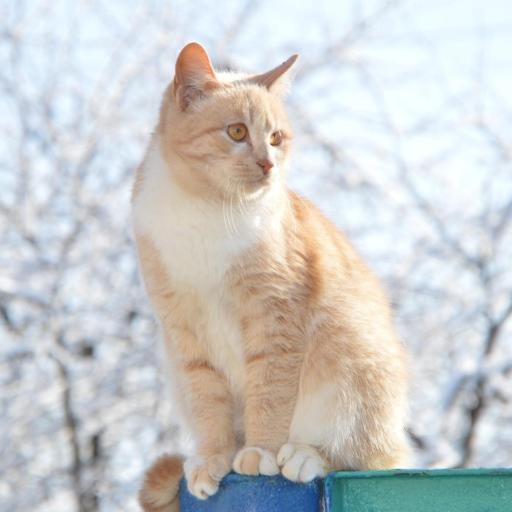} &
    \makecell[b]{A photo of a cat\\wearing a hat\\and glasses.\\~} &
    \includegraphics[width=.09\textwidth]{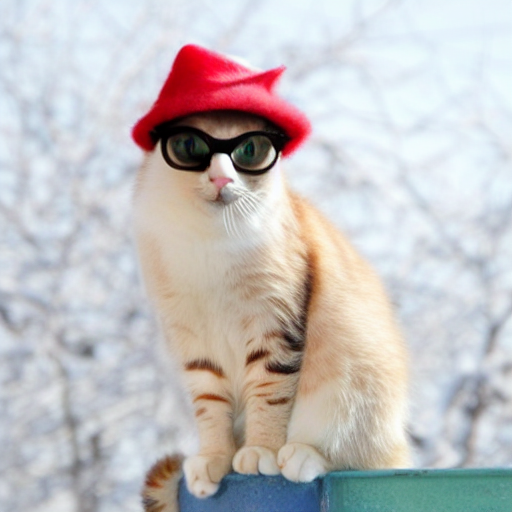} &
    \includegraphics[width=.09\textwidth]{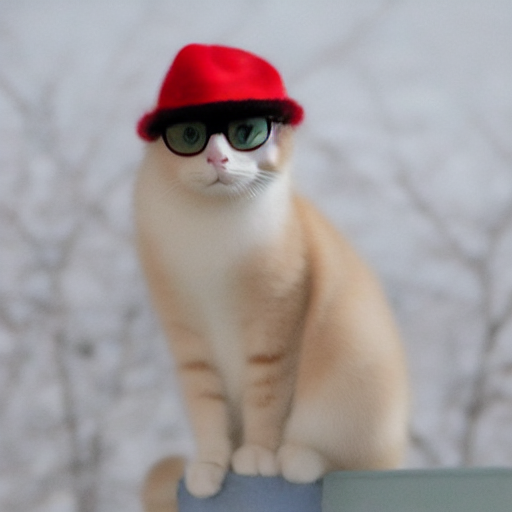} &
    \includegraphics[width=.09\textwidth]{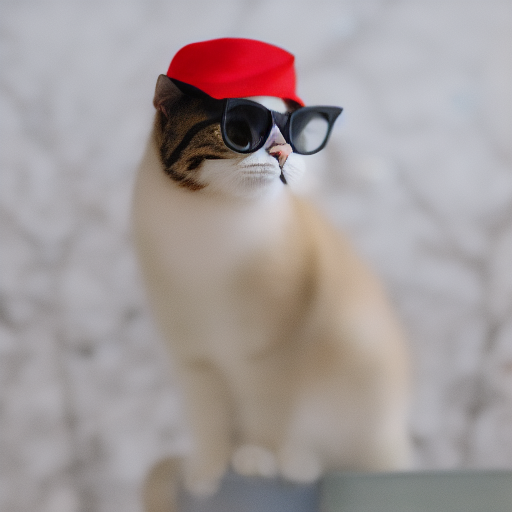} &
    \includegraphics[width=.09\textwidth]{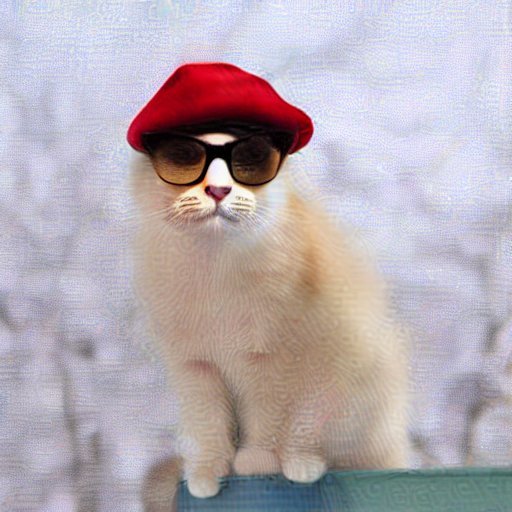} &
    \includegraphics[width=.09\textwidth]{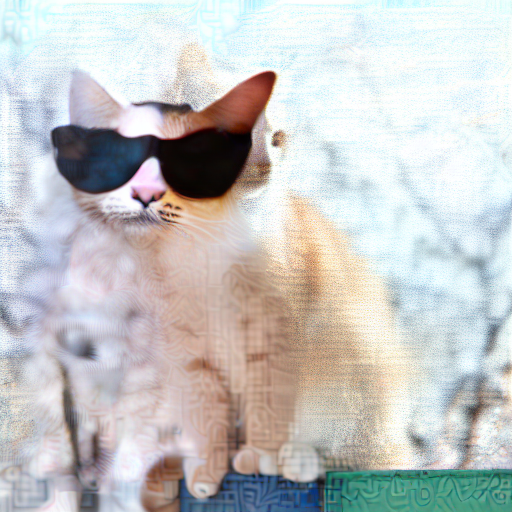} \\

     &
    \includegraphics[width=.09\textwidth]{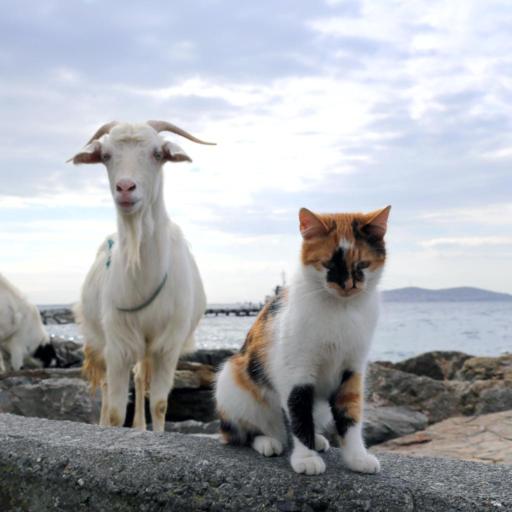} &
    \makecell[b]{A goat to the\\right of a cat.\\~} &
    \includegraphics[width=.09\textwidth]{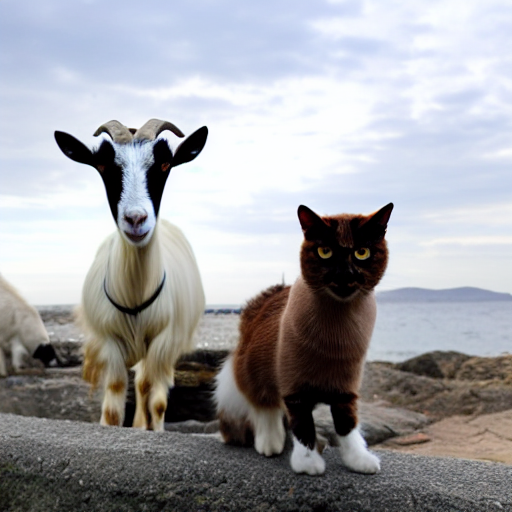} &
    \includegraphics[width=.09\textwidth]{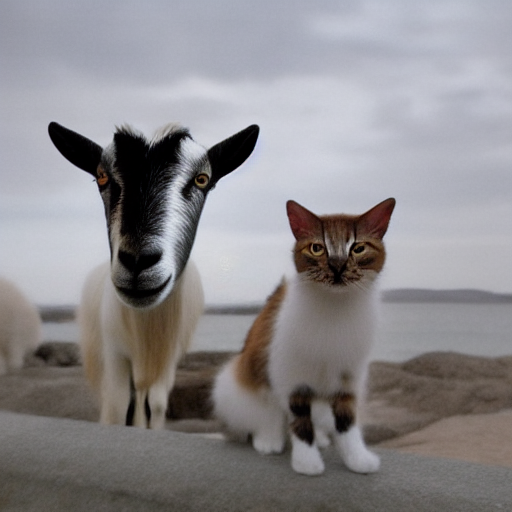} &
    \includegraphics[width=.09\textwidth]{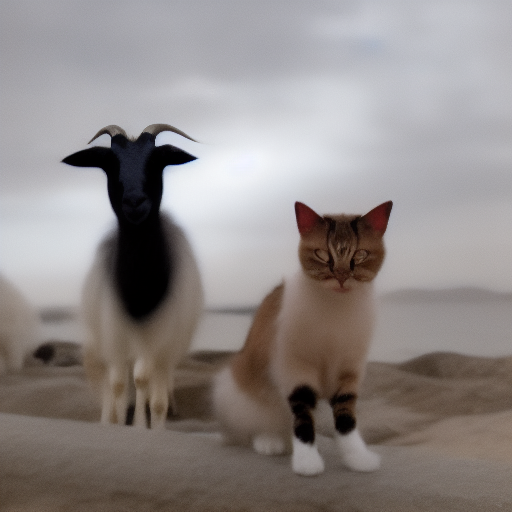} &
    \includegraphics[width=.09\textwidth]{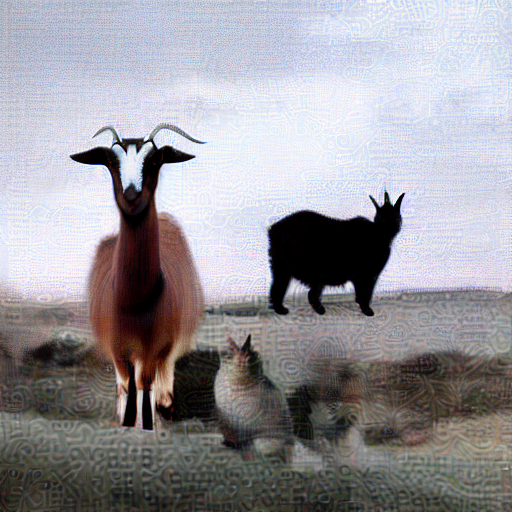} &
    \includegraphics[width=.09\textwidth]{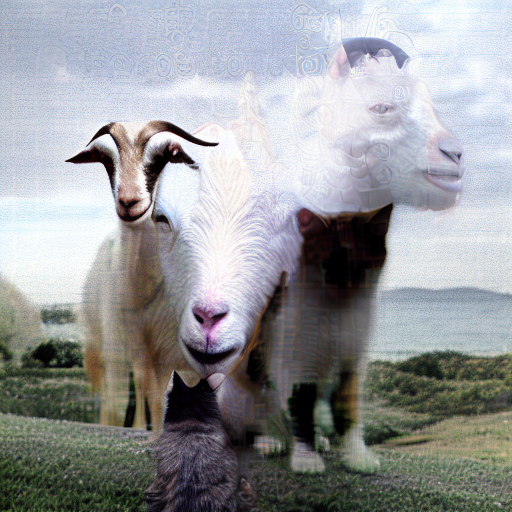} \\

     &
    \includegraphics[width=.09\textwidth]{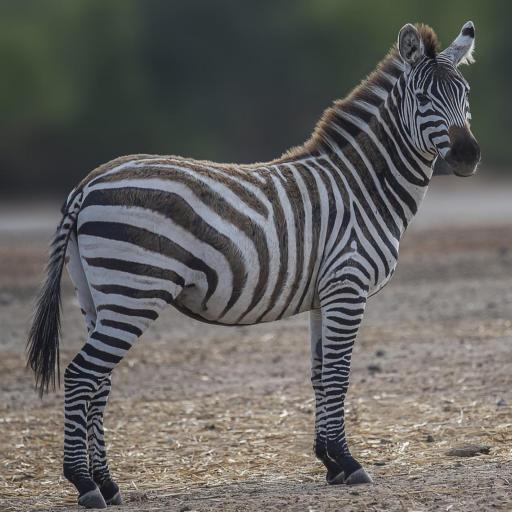} &
    \makecell[b]{A photo of\\a horse.\\~} &
    \includegraphics[width=.09\textwidth]{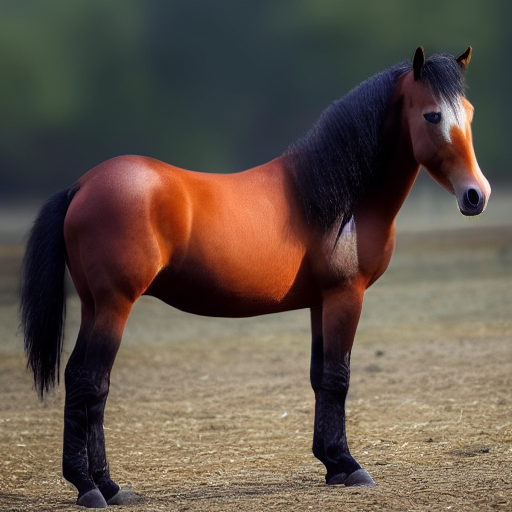} &
    \includegraphics[width=.09\textwidth]{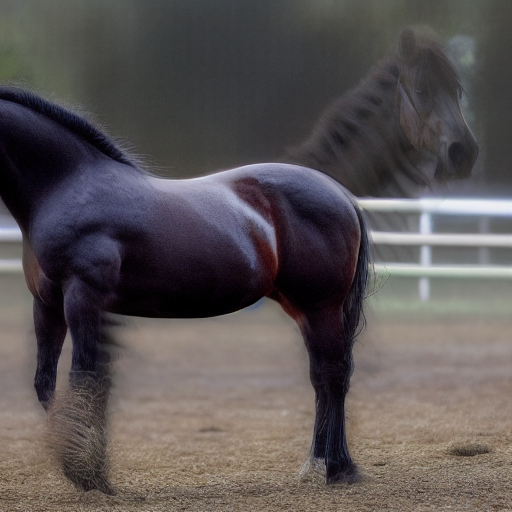} &
    \includegraphics[width=.09\textwidth]{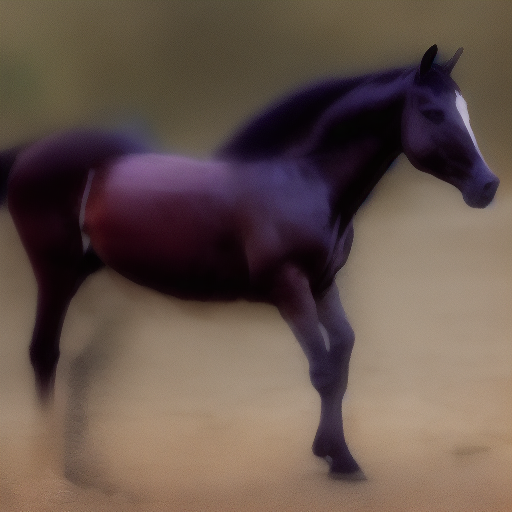} &
    \includegraphics[width=.09\textwidth]{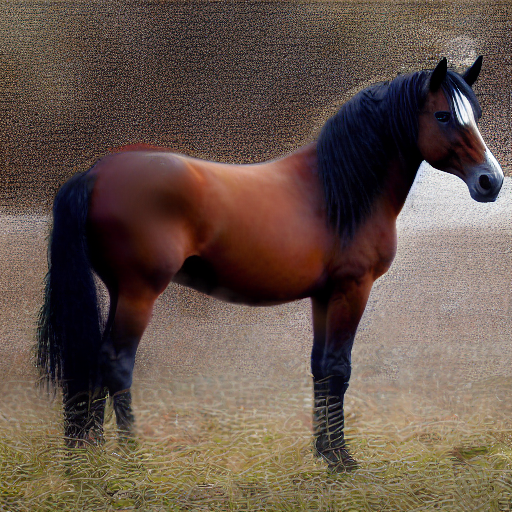} &
    \includegraphics[width=.09\textwidth]{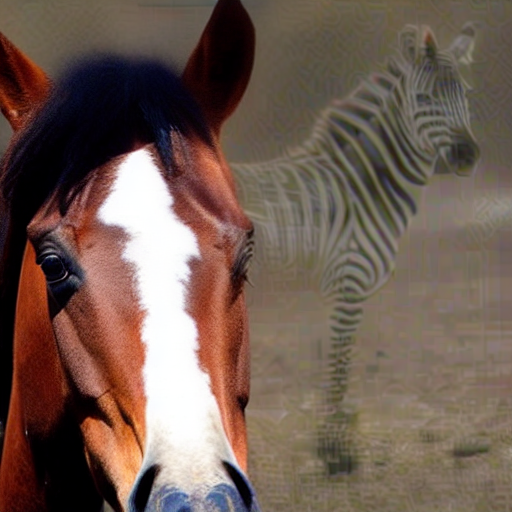} \\ \midrule

    \multirow[b]{3}{*}{SDEdit~\cite{SDEdit}} &
    \includegraphics[width=.09\textwidth]{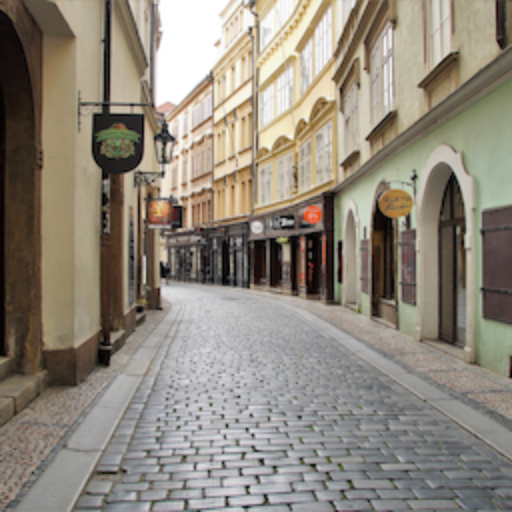} &
    \makecell[b]{A cyclist riding\\in a street.\\~} &
    \includegraphics[width=.09\textwidth]{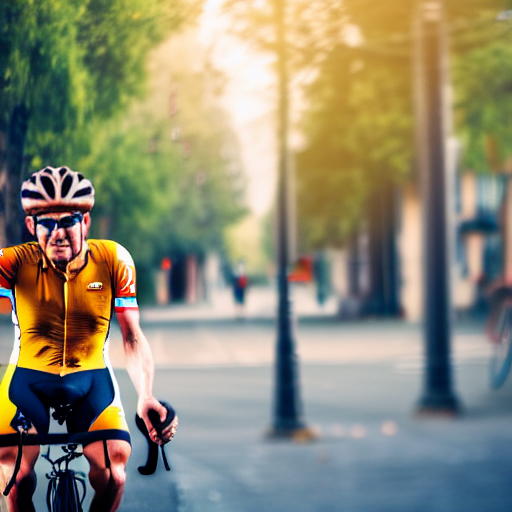} &
    \includegraphics[width=.09\textwidth]{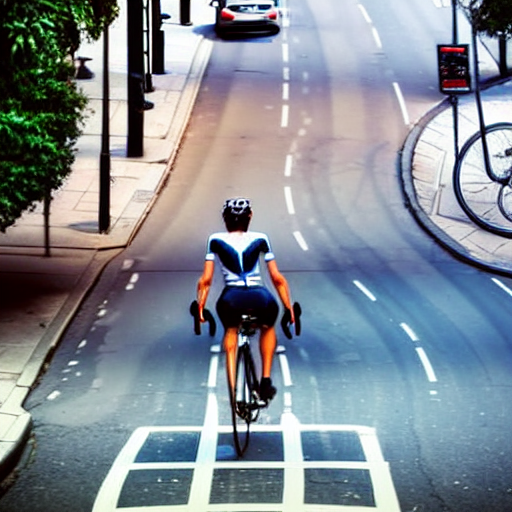} &
    \includegraphics[width=.09\textwidth]{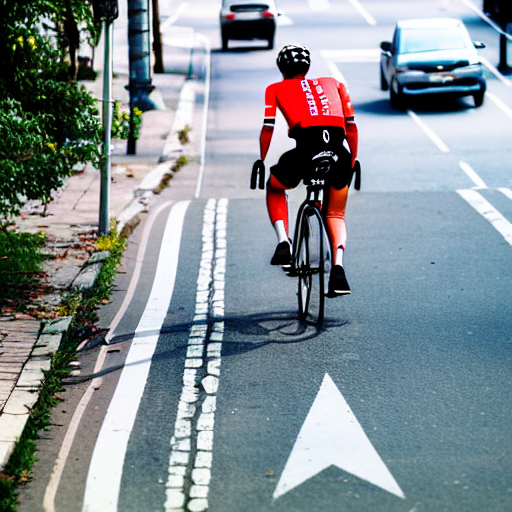} &
    \includegraphics[width=.09\textwidth]{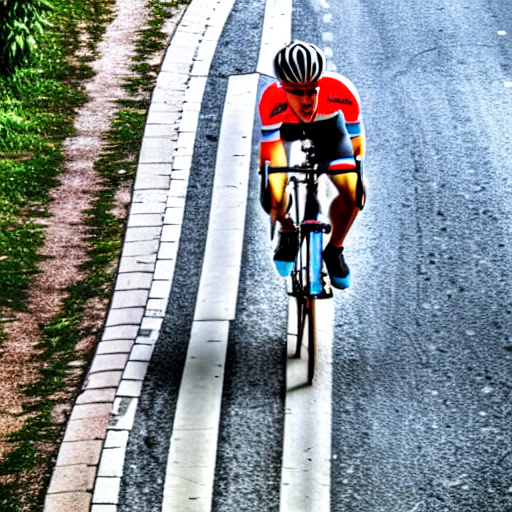} &
    \includegraphics[width=.09\textwidth]{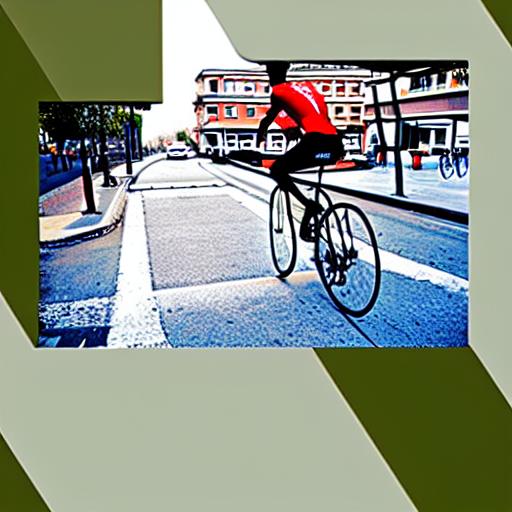} \\

    &
    \includegraphics[width=.09\textwidth]{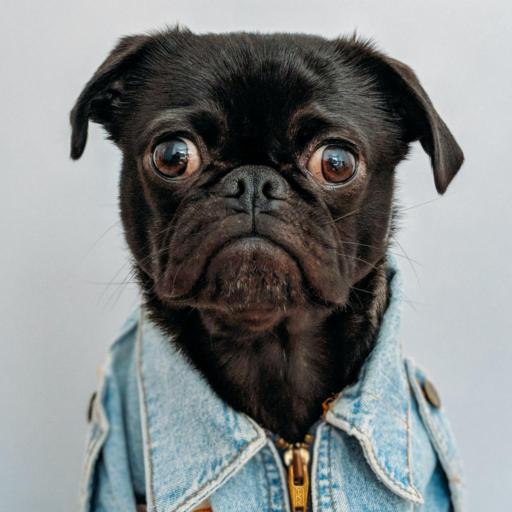} &
    \makecell[b]{A dog smoking\\a cigar.\\~} &
    \includegraphics[width=.09\textwidth]{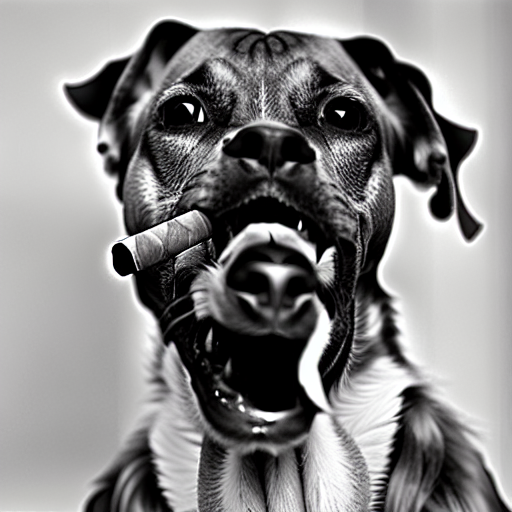} &
    \includegraphics[width=.09\textwidth]{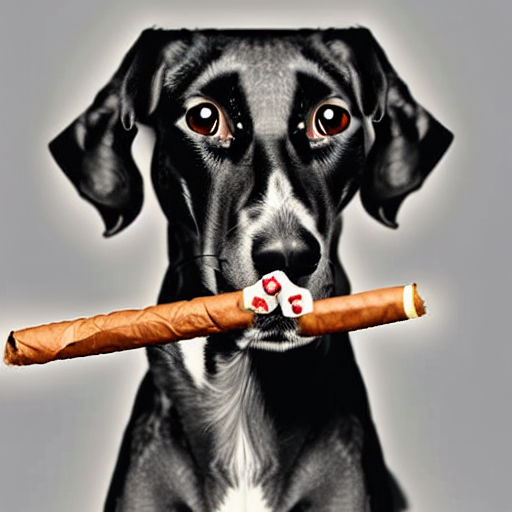} &
    \includegraphics[width=.09\textwidth]{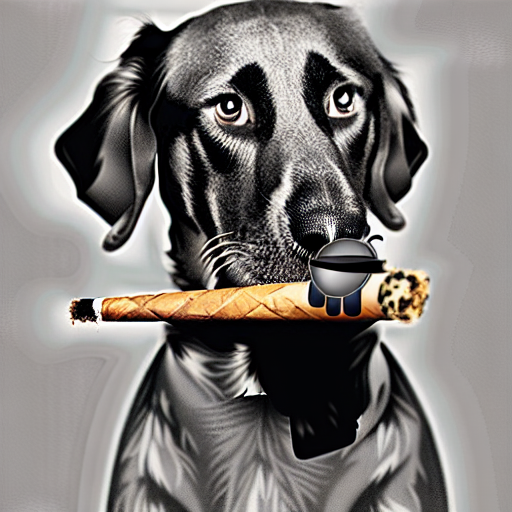} &
    \includegraphics[width=.09\textwidth]{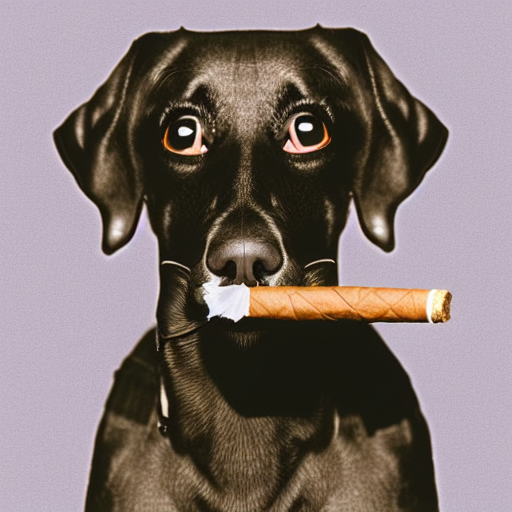} &
    \includegraphics[width=.09\textwidth]{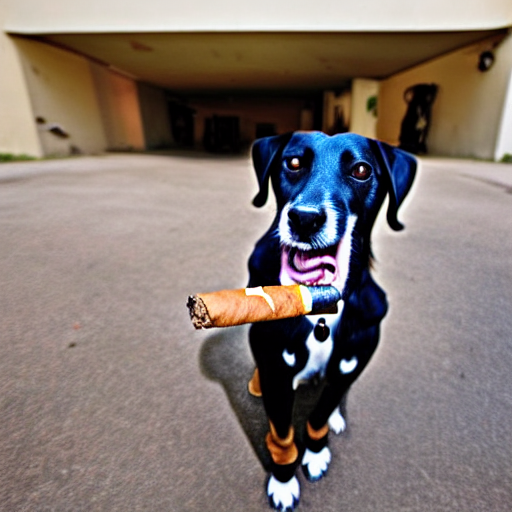} \\

    &
    \includegraphics[width=.09\textwidth]{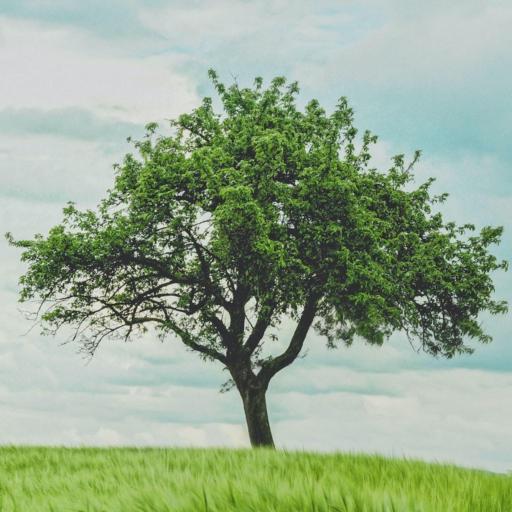} &
    \makecell[b]{A photo of\\a willow tree.\\~} &
    \includegraphics[width=.09\textwidth]{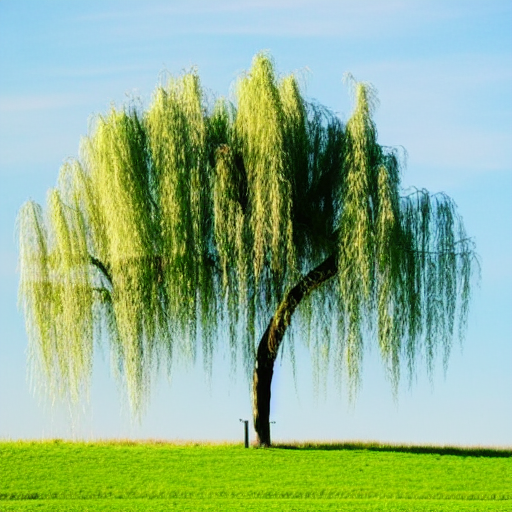} &
    \includegraphics[width=.09\textwidth]{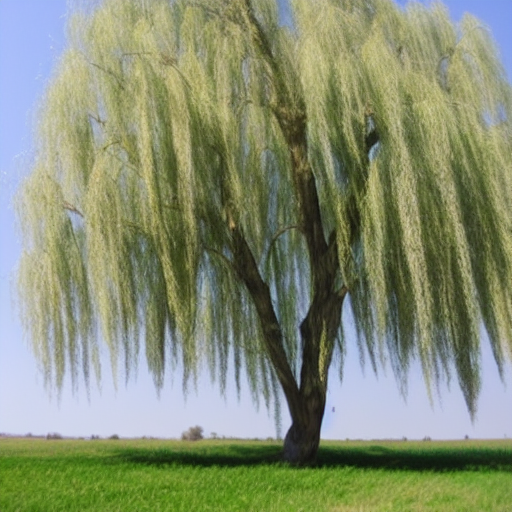} &
    \includegraphics[width=.09\textwidth]{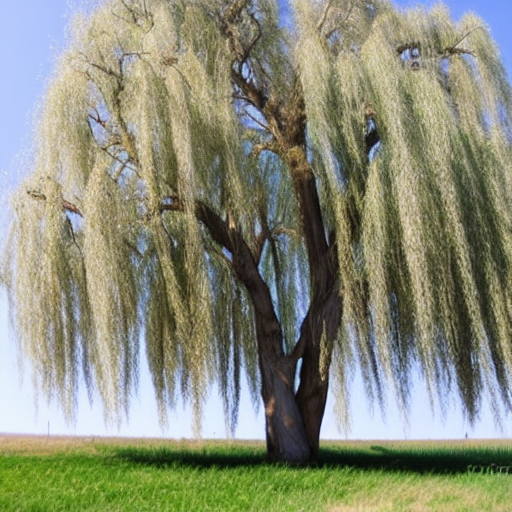} &
    \includegraphics[width=.09\textwidth]{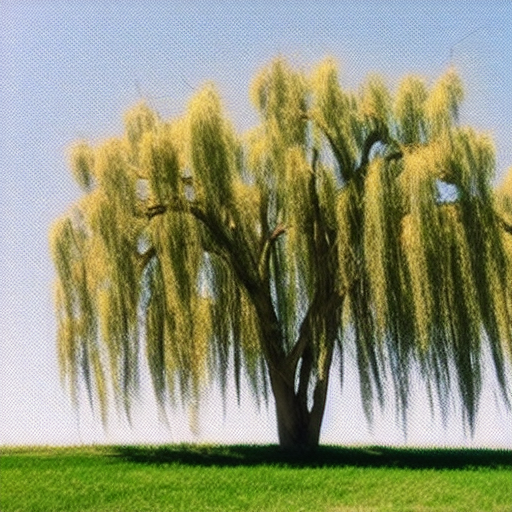} &
    \includegraphics[width=.09\textwidth]{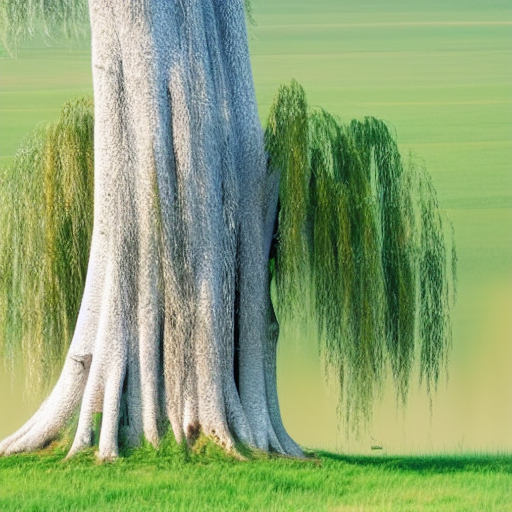} \\
    \bottomrule

\end{tabular}
}
    \caption{Qualitative results. Our attention attack disrupts the spatial layout, resulting more effective than the others.}
    \label{fig:qualitative}
\end{figure*}

A more clear outline is provided by the newly introduced metrics, caption similarity and semantic intersection over union (both oIoU and pIoU). In this case, our method clearly outperforms the baselines consistently. We argue that such metrics better capture the differences between the adversarial and original edits, taking into account both the semantics of the images as well as the spatial layout of the objects they depict.
A qualitative comparison, confirming such insights, is provided in Fig. \ref{fig:qualitative}. It can be clearly seen that the attention attack damages the spatial relations between objects, resulting in an edit that is overall less faithful to the original image.

\subsection{Human Evaluation}
\label{sec:human_eval}
While for classical adversarial machine learning, defining the success of an attack is obvious (e.g., for a classification task, one can simply check whether the output changes), in the context of image editing we encounter some challenges. For instance, an edited image might adhere to the prompt but disregard the input image completely, or vice-versa; or it might comply to the edit while preserving the visual structure of the input while degrading its quality. Finally, it has to be taken into account that, for a given prompt, even the non-perturbed image may not be edited correctly.
Our proposed metrics attempt to measure quantitatively if the spatial layout of the edited image has been disrupted, and how much the attacked image semantically differs from the original edit.

To further verify these aspects, we design a human evaluation procedure asking users to respond to the following four questions. Each question was asked for all three edit algorithms, showing in a randomized sequence the edit results of different immunization approaches and the original edit. Users did not know what edit algorithm was being used nor the immunization approach. The order of the images was randomized for each sample to avoid inducing biases in the annotators.

\textbf{Question 1: Edit Success}. \textit{Select all edits that you would consider successful based on the original image and the prompt. You can select multiple images or no images at all.} Here we ask the user to subjectively express an opinion on the success of the edit. This test will also evaluate the success rate of the edit algorithm.

\textbf{Question 2: Worst Edit}. \textit{Select the edit that you consider to be the worst based on the original image and the prompt. Take into account fidelity to the prompt, similarity to the original image and overall image quality. You can select only one image.} Here, by forcing the user to pick just one image, we reduce the cognitive load for a more complex question and reduce the subjective bias.

\textbf{Question 3: Worst Quality}. \textit{Select the edit with the worst overall quality, ignoring the original image and the prompt. You can select only one image.} Here we disregard the editing process and we only address the quality of the generated image. As before we require a single selection.

\textbf{Question 4: Spatial Layout}. \textit{Select all edits that do not preserve the spatial layout of the original image. Take into account object poses, orientation and presence/absence of other objects. You can select multiple images or no images at all.} In this case we aim to check how attacks are able to alter the overall image structure for both the edited and unedited parts.

In Tab.~\ref{tab:human_tests} we report all results. Q1 shows that  71.3\%  of the edits are considered successful for Ledits++, which is in line with the results reported by the authors \cite{brack2024ledits++}, and have a slightly worse success rate for simpler editing frameworks. Attention attack is always shown to have better performance for all four questions.
In particular, Q4 highlights that our attack largely affects the original layout of the image, especially for DDPM Inversion and LEdits++.

\section{Conclusion}

In this paper, we investigated the effects of attacking cross-attention in diffusion models as an immunization strategy against prompt-based image editing methods. By disrupting the alignment between textual and visual tokens, we can deflect the editing pipeline toward less relevant portions of the image, resulting in edits that do not reflect the original layout and present undesirable visual artifacts. To be agnostic to the prompt, we leveraged image captions as a target for the attack.
To appropriately evaluate the proposed approach we pointed out that common metrics such as PSNR, SSIM and LPIPS are not well suited for editing tasks, as they might not reflect the absence/presence of subtle changes that may make an edit successful or unsuccessful. Similarly, we showed that even the CLIP score metric does not grasp these alterations nor significant changes in object dispositions. We proposed two novel metrics specifically tailored for adversarial edits: Caption Similarity and semantic Intersection over Union (in its two variants, oIoU and pIoU). Our experiments highlight the effectiveness of our approach, which is also validated through human assessment.

\paragraph{Acknowledgements}
This work is partially supported by project "Collaborative Explainable
neuro-symbolic AI for Decision Support Assistant, CAI4DSA, CUP B13C23005640006."

\bibliographystyle{ACM-Reference-Format}
\balance
\bibliography{adversarial_editing}
\end{document}